\documentclass{article}

\usepackage[preprint]{nips_2018}

\usepackage[utf8]{inputenc} % allow utf-8 input
\usepackage[T1]{fontenc}    % use 8-bit T1 fonts
\usepackage{hyperref}       % hyperlinks
\usepackage{url}            % simple URL typesetting
\usepackage{booktabs}       % professional-quality tables
\usepackage{amsfonts}       % blackboard math symbols
\usepackage{nicefrac}       % compact symbols for 1/2, etc.
\usepackage{microtype}      % microtypography

\usepackage{natbib}
\usepackage{setspace}   %Allows double spacing with the \doublespacing command
\usepackage{hyperref}
\hypersetup{
    colorlinks=true,
    linkcolor=blue,
    citecolor=blue,
    filecolor=magenta,      
    urlcolor=cyan
}
\renewcommand{\P}{\mathbb{P}}
\newcommand{\E}{\mathbb{E}}
\newcommand{\R}{\mathbb{R}}

\newcommand{\x}{\mathbf{x}}
\newcommand{\y}{\mathbf{y}}
\newcommand{\z}{\mathbf{z}}

\newcommand{\h}{\mathbf{h}}

\usepackage{graphicx}
\usepackage{amsmath}
\usepackage{subcaption}

\usepackage{color}

\usepackage{lineno}
\allowdisplaybreaks

\usepackage{fancyhdr}
\title{Latent Variable Algorithms \\for Multimodal Learning and Sensor Fusion}

\author{
  Lijiang Guo\\
  Department of Intelligent Systems Engineering\\
  Indiana University\\
  Bloomington, IN 47405 \\
  \texttt{lijguo@indiana.edu}\\
}

\begin{document}
% \nipsfinalcopy is no longer used

\maketitle

\begin{abstract}
%\lijiang{(In version 8-pub, I will remove some parts that I do not want to publish.)}

Multimodal learning has been lacking principled ways of combining information from different modalities and learning a low-dimensional manifold of meaningful representations. We study multimodal learning and sensor fusion from a latent variable perspective. In the first part, we present a regularized recurrent attention filter for sensor fusion. This algorithm can dynamically combine information from different types of sensors in a sequential decision making task. Each sensor is bonded with a modular neural network to maximize utility of its own information. A gating modular neural network dynamically generates a set of mixing weights for outputs from sensor networks by balancing utility of all sensors' information. We design a co-learning mechanism to encourage co-adaption and independent learning of each sensor at the same time, and propose a regularization based co-learning method. In the second part, we focus on recovering the manifold of latent representation. We propose a co-learning approach using probabilistic graphical model which imposes a structural prior on the generative model: multimodal variational RNN (MVRNN) model, and derive a variational lower bound for its objective functions. % and demonstrate in speech processing tasks with multimodal inputs. The algorithm achieves $95+\%$ accuracy on new testing subject with signal to noise ratio of $-5$ DB. 
In the third part, we extend the siamese structure to sensor fusion for robust acoustic event detection. 
We perform experiments to investigate the latent representations that are extracted; works will be done in the following months. Our experiments show that the recurrent attention filter can dynamically combine different sensor inputs according to the information carried in the inputs. We consider MVRNN can identify latent representations that are useful for many downstream tasks such as speech synthesis, activity recognition, and control and planning. Both algorithms are general frameworks which can be applied to other tasks where different types of sensors are jointly used for decision making. To our knowledge, this algorithm is the first that addresses a online multimodal decision making problem. 

%We present Watch And Listen (WAL), a regularized recurrent attention algorithm for sensor fusion which can dynamically combine information from different types of sensors in a online sequential decision making task. Each sensor is bonded with a modular neural network to maximize utility of its own information. A gating modular neural network dynamically generates a set of attention weights for outputs from sensor networks by balancing utility of all sensors' information. We also design a co-learning framework to encourage co-adaption and independent learning of each sensor at the same time. We first proposal a regularization based co-learning. Then we present a novel co-learning approach using structured probabilistic graphical model and multimodal stochastic RNN (MSRNN). We derive a variational lower bound for the MSRNN model objective function. We demonstrate our algorithm in speech processing tasks with multimodal inputs. %The algorithm achieves $95+\%$ accuracy on new testing subject with signal to noise ratio of $-5$ DB. 
%Our experiments show that the WAL algorithm can dynamically combine different sensor inputs according to the information carried in the input. WAL algorithm is a general framework which can be applied to other tasks where different types of sensors are jointly used for decision making. To our knowledge, this algorithm is the first that addresses a online multimodal decision making problem.
\end{abstract}

% Create hyterlink to Table of Contents
\hypertarget{ToC}{}
\newpage
\tableofcontents
\newpage
% Add header and footer to each page.
% Add fancy header and footer
% \newcommand{\courseno}{S626 Project}
% \lhead{\textbf{\courseno}}
% \chead{\textbf{Lijiang Guo}}
% \rhead{\textbf{Updated \today}}
\rhead{\hyperlink{ToC}{Contents}}

% \section{Note}
% We combine 
% \begin{enumerate}
%     \item recurrent attention
%     \item mixture of local experts
%     \item variation of information
%     \item regularization
%     \item variational inference
%     \item onine learning
% \end{enumerate}

\section{Executive Summary}
We present three multimodal learning and sensor fusion methods from a latent variable perspective. In \autoref{sec:introduction} we briefly introduce the multimodal learning problem, \autoref{related-works-multimodal-learning} reviews background with concrete examples. In \autoref{sec-model-agnostic-fusion}, \autoref{sec-msrnn}, and \autoref{sec:siamese} we introduce three different methods: a model-free sensor fusion approach, a model-based sensor fusion approach, and sensor fusion for weakly supervised embedding. In each of these sections we first discuss related works, then present our approach. Experiment results are discussed in \autoref{sec:experiment}. We discuss future works in \autoref{sec:future}.

\section{Introduction}\label{sec:introduction}
Multimodal learning has been studied in various forms for decades. Broadly speaking, multimodal input can be considered as streams of loosely synchronized multi-domensional data, with each modality being one subset of dimensions. In simple cases when measurements are both taken in real domain, with certain metrics assumed, we often impose a model which leverages on the dependency across modalities to jointly solve a problem. Two commonly used dependency models are linear correlation and conditional probability. 

Consider a example from \citet{shumway2011time} where a patient's biometric markers such as log(white blood count) [WBC], log(platelet) [PLT], and hematocrit [HCT] are used to predict probability of a patient's long term survival using Bernoulli linear regression:
\begin{linenomath}
\begin{align}
    \E[Y|X] = f(\beta_0 + X_\text{WBC}\beta_1 + X_\text{PLT}\beta_2 + X_\text{HCT}\beta_3),
\end{align}
\end{linenomath}
where $f$ is a proper transformation function. We can consider $X_\text{WBC}, X_\text{PLT}, X_\text{HCT}$ as different input modalities as they are essentially measuring different physical concepts. Classical statistical solutions focus on statistical inference in the sense of quantifying the error in estimating $\E[Y|X]$ using probability distributions. To this end, (1) $f$ is often chosen to be functions which are friendly to manipulate, and (2) the relation between different modalities are chosen to be simple linear additive for the same reason. In this simple linear model for regression, we use linear combinations of the input variables. We can extend this model by considering linear combinations of fixed nonlinear functions of the input variables:
\begin{linenomath}
\begin{align}
    \E[Y|x] = f\left(\beta_0 + \phi_1(X_1)\beta_1 + ... + \phi_k(X_K)\beta_K \right).
\end{align}
\end{linenomath}
In this model, the relation between different input variables are still linear additive, but we allow for flexible representation, i.e. $\phi$, of each variable, i.e. features, to be learned from training by optimizing some criterion. 

Different from classical statistics, in machine learning we often face very high dimensional input data, and each input modality has essentially different representation forms; for example, video, audio, and text. The challenging questions in creating intelligent systems for processing audio or video inputs are
\begin{enumerate}
    \item How to describe the complex relation between a single input with the target output, and
    \item How to combine different input modalities when there are no straight forward physical model to describe them.
\end{enumerate}
Let's consider speech recognition as a example. Consider we have two input modalities, audio and video, and text as output. The first question corresponds to how are we going to describe the relation between audio and text, or video and text, respectively. The second question corresponds to how the audio input is related to video input in terms of text generation. In this research, we address the second question in a principled manner, and, equally important, create a multimodal algorithm that can outperform uni-modal algorithms.

We want to focus on multimodal time-series data as they are universal in speech, video, computer network, robot control, etc. Our objective is sequential inference under sequential multimodal input. Consider we have two input streams $\{X_t\}$ and $\{Y_t\}$, and a output steam $\{L_t\}$ where $t \in \{1,.., N\}$ are time indices. Our objective is to predict $L_{k}$ from $\{(X_t,Y_t)\}_{t=1}^{k}$. The challenge is to model $X_t$ given ($X_{t-1}, ..., X_1, Y_t, Y_{t-1}, ..., Y_1)$, and to model $Y_t$ given ($Y_{t-1}, ..., Y_1, X_t, X_{t-1}, ..., X_1)$. This is particularly important when one input modality is corrupted, and the multimodal algorithm has to recognize or infer the missing information from other input modalities. In particular, we want a algorithm which can estimate the reliability of each sensor input. We will discuss this in detail in \autoref{sec-model-agnostic-fusion}.

\begin{figure}[t!]
    \centering
    \includegraphics[width=0.3\textwidth]{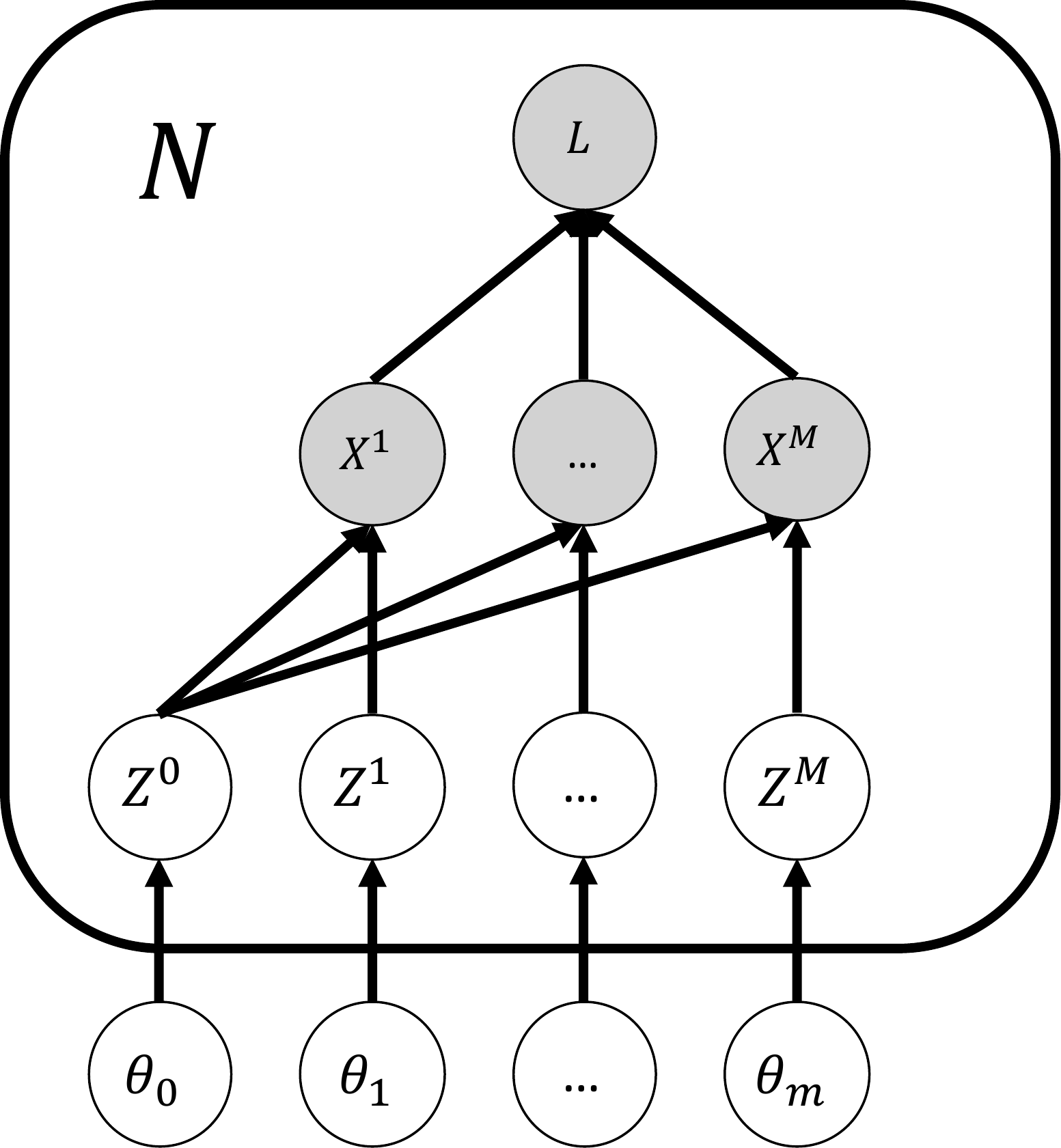}
    \caption{Structural generative model.}
    \label{fig:GM}
\end{figure}

From a generative model perspective, we assume that there are some latent random variable $Z$ which generates the observed random variables $X$ (see \autoref{fig:GM}). In multimodal data, $X$ can be partitioned into disjoint sets of input signal where each  $X^{(m)}$ corresponds to one modality. The observed speech activity state $L$ is a function of $X$. The latent variables $Z$ can also be partitioned into modality invariant and modality dependent subsets $\{Z^{(0)},Z^{(1)},...,Z^{(k)} \}$ where $Z^{(0)}$ is shared by all modality $\{X^{(m)}\}_{m=1}^k$, and each $Z^{(m)}$ is specific to $X^{(m)}$ for $m\in\{1,...,k\}$. Having a shared latent variable $Z^{(0)}$ enables the model to explain for the synchronization between different modalities. In \autoref{sec-msrnn} we discuss this in detail.

% A conditional generative model is more flexible to model a distribution of multiple modes \cite{sohn2015learning, kingma2014semi}. In addition to having a generative model, the conditional generative allows for a discriminative prediction.
%
% \begin{figure}[t!]
%     \centering
%     \begin{subfigure}[b]{0.3\textwidth}
%         \includegraphics[width=\textwidth]{figures/GM.pdf}
%         \caption{Regular generative model}
%         \label{fig:GM}
%     \end{subfigure}
%     \qquad
%     %~ %add desired spacing between images, e. g. ~, \quad, \qquad, \hfill etc. 
%       %(or a blank line to force the subfigure onto a new line)
%     \begin{subfigure}[b]{0.37\textwidth}
%         \includegraphics[width=\textwidth]{figures/CGM.pdf}
%         \caption{Conditional generative model}
%         \label{fig:CGM}
%     \end{subfigure}
%     \caption{Graphical model representation of generative and conditional generative models.}\label{fig:animals}
% \end{figure}

We will begin by reviewing a few related topics. Then we present a deep learning based regularized recurrent attention filter, which dynamically combines information from multiple sensor inputs in a sequential decision making task. We present a co-learning mechanism and discuss its advantages in learning robust latent features. Next we marry the strength of deep learning with probablistic graphical models in a coherent way for multimodal learning. We present a multimodal nonlinear state-space model with a structural prior for separating modality dependent and modality invariant features. To demonstrate the proposed algorithms, we tackle the problem of speech activity detection and speech separation in corrupted audio-video streams.

This paper is a first step towards a faraway goal, thus we cannot solve the problem completely. Our model has the following contributions (no. 3, 4 and 5 are unique to our knowledge):
\begin{enumerate}
    \item The model combines information collected from multiple sensors. Each sensor collects different types of data; for example, image, motion, audio, etc.
    \item The model has independent processing module for each sensor. Each module is developed to maximize the utility of its sensor input in a decision making task.
    \item The model addresses a sequential decision making task. It (1) takes new inputs at each time step, and (2) generates new attention over modalities by combining the new inputs and its memory, then (3) apply attention to modules to make a final decision for the current time step. The model does not need future information to make decision for current time step. %To our knowledge, we are the first to propose an algorithm that address a online multimodal learning problem using attention.
    \item Each module is divided into two sub-components. One sub-component co-learns with other modules; this sub-component allows for simultaneous co-learning features that are shared by all sensors, which allows for robust detection of these features, even a subset of sensors' inputs are corrupted. The other sub-component learns independently from other modules; this allows each module to focus on its sensor's unique input features that are not shared by other sensors. This co-learning design prevents overfitting during training, and improves robustness of the modules.
    \item The probablistic approach to co-learning can be used to separate modality-dependent features from modality invariant features. It has wide applications such as speech recognition, speaker identity recognition, etc.
\end{enumerate}

\section{Multimodal Learning and Sensor Fusion}\label{related-works-multimodal-learning}
\subsection{An Example: Audio-Visual Learning}
Human senses the world with multiple modalities such as vision, sound, texture, etc. An AI agent, e.g. a robot, also relies on multiple sensors to collect data to perform tasks such as localization, path planning, control a robotic arm, etc. \citet{baltruvsaitis2018multimodal} refers a \emph{sensory modality} as our primary channels of communication and sensation, such as vision or touch. A problem is characterized as \emph{multimodal} when it includes multiple sensory modalities. In this work, as a concrete example, we focus on three modalities that are important to human speech: visual, audio, and motion. However, the method we discuss are general enough to tackle other sensor modalities on AI agents. 

Many researches leverage natural synchrony between simultaneously recorded visual and audio signals to solve speech related problems. In recent works, lip-reading, the practice of using visual signals to understand speech, has been proven to be able to effectively extract useful information for speech recognition \citep{ephrat2017improved, assael2016lipnet}. Hence it is natural to consider audio-visual approaches for many speech related tasks, including speech recognition \citep{ngiam2011multimodal, mroueh2015deep, feng2017audio}, speech separation \citep{hou2017audio}, and voice activity detection \citep{ariav2018deep}.%, and speaker identification \citep{ren2016look}, and voice activity detection \citep{kahou2016emonets}.

Among these problems, speech separation is one of the fundamental problems in audio processing. \citet{wang2018supervised} gave a overview of recent advances in deep-learning based audio-only speech separation system. Traditionally audio-only methods are used to separate a speech signal from other background signals. When multiple human speeches overlap each other, the speech separation problem is referred to as the \emph{cocktail party problem}. Such a problem is especially challenging to solve because human speeches tend to have similar features than with non-speech sounds. It is also difficult to assign a speech to the corresponding speaker with audio signals only, often referred as \emph{label permutation problem}. A few solutions have been proposed to address multi-speaker speech separation using single channel audio recording. For example, \citet{hershey2016deep} proposed a method called deep clustering (DPCL) which uses discriminatively trained speech embeddings to cluster and separate speeches. They also proposed a permutation invariant loss function to solve the label permutation problem. \cite{yu2017permutation} and \cite{isik2016single} successfully use a permutation invariant loss function to train a DNN for multi-speaker speech separation. 

In recent years, deep learning based audio-visual methods have been used for speech separation. \citet{hou2017audio} proposed a CNN-based model which outputs a denoised speech spectrogram as well as a reconstruction of the input mouth region. \citet{gabbay2017visual} trained a speech enhancement model on videos where other speech samples of the same speaker are used as background noise. \citet{gabbay2018seeing} use a video-to-sound synthesis method to filter noisy audio. While most audio-visual speech separation methods are speaker dependent, \citet{ephrat2018looking} proposed a speaker-independent audio-visual model for separating single speech from a mixture of sounds such as multiple speakers and background noise. They use a pretrained face detection model to identify all speakers in a video, then outputs a complex spectrogram mask for each speaker. %\cite{afouras2018conversation} propose a similar model where a phase network and a magnitude network work jointly to separate a speech from the background. 
Alternatively, \citet{owens2018audio} train a deep neural network to predict whether audio and visual streams are temporally aligned. Learned features extracted from this self-supervised model are then used to condition an on/off screen speakers source separation model. \citet{gao2018learning} and \citet{zhao2018sound} addressed the closely related problem of separating the sound of multiple on-screen objects (e.g. musical instruments). 

%Audio-visual methods have also been used to solve other speech related problems. In a speaker identification task, \citet{ren2016look} proposed a multimodal LSTM architecture which shares weights across time steps as well as across modalities. They showed that modeling the temporal dependency across face and voice can improve the robustness to content quality degradations and variations. 

\subsection{An Example: Load Balancing in Distributed System for Streaming Data}
Suppose we have $K$ streams of input data $x^{(1)}, ..., x^{(K)}$, each are different but correlated. They are jointly used to solve a problem whose answer is a stream $y$. Consider we have many \emph{workers} $\{w^{(1)},..., w^{(N)}\}$ who are processing these inputs $\{x_k\}_1^K$. In a distributed system we need to partition the input streams to assign to different workers, and then gather them to solve for the answer. Let's assume for stream $k$, the information complexity fluctuates in time according to an unknown function $h^{(k)}(x)$, such that the processing time of a worker $w$ is proportional to $t(k, x) = f(h^{(k)}(x))$. For example, a encoded video could take various amount of computations to decode. 

%Is there a way for us to predict $h^{(k)}(x)$ such that we can assign workers according to information complexity?

%\subsubsection{Computer Network}
%\subsubsection{CPU Load Balancing}

\subsection{Sensor Fusion}
Sensor fusion is one of the core problems in multimodal learning. Technically speaking, in multimodal sensor fusion, we need to integrate information from multiple modalities with the goal of fulfilling some tasks. Comparing with unimodal methods, \citet{baltruvsaitis2018multimodal} suggested multimodal sensor fusion has the benefit of (1) providing more robust predictions, (2) allowing the model to utilize complementary information from different modalities, and (3) imputing missing information for corrupted signals. 

Sensor fusion algorithms can be divided into two big categories \citep{baltruvsaitis2018multimodal}: model-agnostic and model-based. Model-agnostic sensor fusion does not depend on a specific machine learning method. It usually combines different sensors in a blind way and the relationship between sensors are not exploited. Model-based methods explicitly address fusion in their construction, e.g. using graphical models to impose a structural prior on modalities. In this work, we will propose both a model-agnostic approach and a model-based approach for sensor fusion.

\begin{figure}[t!]
    \centering
    \includegraphics[width=0.4\textwidth]{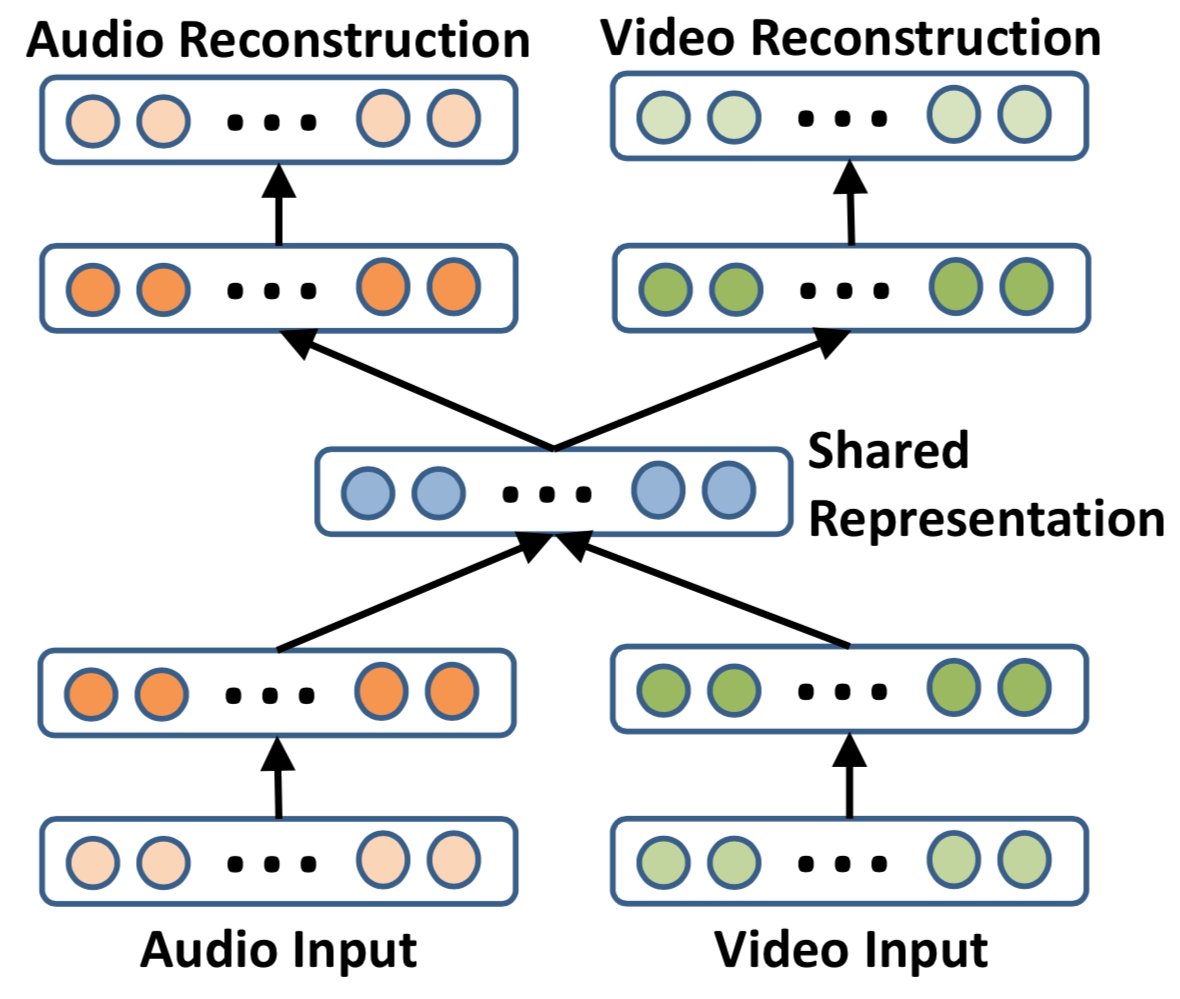}
    \caption{Multimodal deep Boltzmann machine (reproduced from \cite{ngiam2011multimodal}).}
    \label{fig:ngiam-multimodal-DBM}
\end{figure}

The most common fusion approach is to concatenate the inputs from different sensors at certain modeling stage to solve a unimodal learning problem. \cite{ngiam2011multimodal} proposed a multimodal autoencoder which first use deep Boltzmann machine (DBM) to learn each audio and video features independently, then concatenate the learned features to an multimodal autoencoder to learn a shared representation (\autoref{fig:ngiam-multimodal-DBM}). To reproduce original inputs, the shared representation is mapped back to each modality. \cite{srivastava2012multimodal} also use DBM to learn a generative model of multimodal input data. \cite{simonyan2014two} proposed a two-steam convolutional neural networks for action recognition in video. The two-stream model try to capture the complementary information from two modalities: still image and motion between frames. In their work, the last layers concatenates the two streams into a final classification layer. \cite{torfi20173d} proposed a coupled 3D-CNN to map multiple modalities into a representation space to evaluate the correspondence of audio-visual streams. In \citet{ephrat2018looking} the audio and visual streams are combined by concatenating the feature maps of each stream, which are subsequently fed into a BLSTM followed by three FC layers.

Whether to combine information at early or late stage affects sensor fusion results. %Because it is much harder to capture relationship across modalities than features within a modality, 
\cite{ngiam2011multimodal} argued that naive concatenation of different sensors' inputs discourages co-learning since the within sensor correlation is much stronger than between sensor correlation, hence the model are biased towards learning dominant patterns in each sensor separately instead of learning patterns that occur simultaneously in multiple sensors. In this regard, they propose to learn joint representations that are shared across multiple modalities at the higher layer of the deep network after learning layers of modality-specific networks. As \cite{sohn2014improved} commented, the rationale is that, they believe, the learned features may have less within-modality correlation than raw features, and this makes it easier to capture patterns across data modalities. However, in \cite{ngiam2011multimodal}, the middle layer is a simple fully connected layer, and we can only hope that it can always capture meaningful representations shared by all modalities. %\lijiang{(May add definition of early and late fusion as in \citep{baltruvsaitis2018multimodal} and  discussion about early and late fusion from \cite{hori2017early})}

%Another way of model-agnostic fusion is through weight sharing. In a speaker identification task, \citet{ren2016look} proposed a multimodal LSTM architecture which shares weights across time steps as well as across modalities. They showed that modeling the temporal dependency across face and voice can improve the robustness to content quality degradations and variations.

One challenge of multimodal learning is how to deal with missing data in different modalities, and infer the missing data from other modalities --- it is a unique advantage of multimodal learning over unimodal learning as we can use the information extracted from one modality to improve the recognition ability of the other modality by complementing the missing information, i.e. given two input streams $\{X\}$ and $\{Y\}$, how to model $X_t$ given ($X_{t-1}, ..., X_1, Y_t, Y_{t-1}, ..., Y_1)$, and $Y_t$ given ($Y_{t-1}, ..., Y_1, X_t, X_{t-1}, ..., X_1)$. A naive approach is to use a autoencoder to map all modalities into a common latent space, and then map back to each modality, hoping that the latent representation can recover all modalities. For example, \cite{ngiam2011multimodal} and \cite{ariav2018deep} use a sparse autoencoder to map different modalities into the same space. \cite{ngiam2011multimodal} requires each pre-trained modality-specific network is also jointly trained with other modality-specific networks for cross-modality learning. To infer missing data in one modality, they suppress other modalities with zero input during training.

Model-based sensor fusion explicitly incorporate the relation between sensors into the fusion stage. For example, we can combine multiple sensor signals into modality-dependent and modality-invariant latent random variables through a probabilistic graphical model. Model-based sensor fusion has the advantage of probabilistically detect and recover corrupted signal. We will discuss model-based sensor fusion in detail in \autoref{sec-msrnn}.

In a interesting work, \cite{sohn2014improved} proposed a multimodal representation learning framework which minimizes the information distance between data modalities through the shared latent representations. To this end, they use variation of information to measure the conditional probability of each modality given the other modalities, thereby determine the quality of shared representation across modalities. As a generative model it can predict missing data modalities given partial observation.

Despite the promise, there still remains missing a principled way of how to learn a good association between multiple data modalities that can effectively deal with missing data modalities in the testing time. Both \cite{ngiam2011multimodal} and \cite{sohn2014improved}'s approaches promote learning a shared representation by different modalities, and rely on the collaboration of different modalities. However, co-adaption of different modalities may reduce the individual strength of each modality as we discuss in next sections. We propose two strategies to coordinate the collaboration between modalities. In the first approach we let each local expert focus on it's own modality, and let a dedicated gate expert to coordinate experts. We will discuss this in \autoref{sec-model-agnostic-fusion}. In the second approach, we propose a probabilistic graphical model which decompose latent representation into modality dependent and modality invariant features. We will discuss this in \autoref{sec-msrnn}. 

% A different stream of multimodal learning focuses on a discriminative model approach. These model uses various kinds of strategies to 

% Shared representation. Similarity in the representation space implies similarity of the corresponding concepts from apparently different input modalities. 

% Naive reconstruction loss \cite{ngiam2011multimodal}\cite{srivastava2012multimodal}.

% Use DBM to learn a joint density model  over the space of multimodal inputs \cite{srivastava2014dropout}. This has the advantage that missing modalities can then be filled-in by sampling from the conditional distributions over them given the observed ones.

\section{Model Combination for Sensor Fusion}\label{sec-model-agnostic-fusion}
\subsection{Motivation}
In this section we will develop a model-agnostic sensor fusion algorithm. We are motivated by model combination and aggregation as a way to use parallel models to jointly solve a common problem. Model combination algorithms such as \emph{boosting} has been proven to be a effective approach in reducing prediction error and variance \citep{bishop2006pattern}. A natural way to approach multimodal learning is to divide a task into multiple parallel sub-tasks such that each modality uses a dedicated processing module to solve a sub-task. We consider a \emph{conditional mixture model} \citep{bishop2006pattern} to coordinate all processing modules towards a common learning goal. From the success of \emph{attention mechanism} in sequence-to-sequence model, we observe that mixing weights can be generated in analogous to a spatial attention through a deep neural network. Therefor we combine model combination methods with recent development in deep learning to propose a recurrent attention filter for multimodal sensor fusion. %\lijiang{(Need more revision.)}

%Tree based models, 
%
%Mixture of experts (Jacobs, et. al. 1991). 
%
%EM with iterative re-weighted least square solution  (Jordan and Jacobs, 1994)
%
%Hierarchical mixture of experts (Jordan and Jacobs, 1994).
%
%A Bayesian treatment of HME (Bishop and Svensen, 2003)
%
%Mixture density network (Bishop, 1994)
%
%When the mixture coefficients are input dependent, the hierarchical model becomes nontrivial. 
%
%Attention

\subsection{Related Works}

\subsubsection{Temporal Attention}
In sequential prediction problems, sequence-to-sequence (seq2seq) model is a ground-breaking work \citep{sutskever2014sequence,  cho2014learning}. One problem with seq2seq networks is that their performance will deteriorate rapidly as the length of input sequence increases \citep{cho2014properties}. RNN such as LSTM still suffers from difficulty in memorizing long-range relations, especially when a lot of information is cluttered in a sequence. The recently proposed attention mechanism solved this problem by providing a skip-connection and let the decoder focus on a sub-region of the input \citep{bahdanau2014neural, chorowski2015attention, mnih2014recurrent, ba2014multiple}. Instead of memorizing the entire information of the past sequence, attention filter let the RNN memorize the location of the past information. This greatly reduces the information load of the RNN memory cell. Recently, convolutional and fully-attentional feed-forward architectures like the Transformer model \citep{vaswani2017attention} have emerged as a viable alternative to RNNs for a range of sequence modeling tasks.

In \citep{bahdanau2014neural}, attention mechanism was introduced to help the decoder RNN to focus on relevant parts of the input sequence without relying on the RNN to encode all the information through repeated updating the hidden unit. Recall in a seq2seq model, the encoder takes a length $M$ input sequence $\{x_{t}\}_{t=1}^m$ and generates a \emph{encoder} hidden state sequence $\{h^e_{t}\}_{t=1}^M$. The last hidden state of encoder $h^e_M$ is used as the initial hidden state for the decoder $h^d_0$ to generate another state sequence $\{h^d_{t}\}_{t=1}^N$. During training, to making the decoder learn faster, the decoder also takes a auxiliary input which is the true label sequence, $\{y_t\}_{t=1}^N$; during testing, use decoder output prediction $y_{t-1}$ as input to step $t$. In this simple encoder-decoder structure, suppose the last hidden node of the decoder, $h^d_N$ is related to the first hidden node of the encoder, $h^e_1$, then the relevant information has to pass through $M+N$ RNN transitions. In general, the average distance between encoder and decoder hidden units is $O(M+N)$ --- the number of operations required to relate signals from two arbitrary input or output positions grows linearly in the distance between positions. This long path has made if more difficult to learn dependencies between distant positions. RNN such as LSTM suffers from difficulty in memorizing long-range relations, especially when a lot of information is cluttered in a sequence. To solve this problem, \citet{bahdanau2014neural} proposed attention mechanism which provides a skip-connection and let the decoder focus on a sub-region of the input. Instead of memorizing the entire information of the past sequence, attention filter let the RNN remember the location of the past information. This greatly reduce the information load of the RNN. With attention, the average distance between encoder and decoder hidden units is $O(1)$ --- a constant number of operations.

%\lijiang{(I wonder if I should remove the following technical explanations of attention mechanism, as they are mostly from literature.)}
The attention mechanism is similar to searching in a database and involves a interaction between query, key, and values. Consider the input to time step $t$ of the decoder as a \textbf{query} (i.e. input is the previous hidden state $h^d_{t-1}$), 
\begin{linenomath}
\begin{align}
    q_t =h^d_{t-1}.
\end{align}
\end{linenomath}
The encoder hidden states $\{h^e_t\}_{t=1}^M$ are \textbf{keys} whose information are the \textbf{values}. For example, $h^e_t$ is the key to $t$-th encoder state. To get the $t$-th decoder output $y_t$, we search for the relevant encoders values by comparing its key with the query,
\begin{linenomath}
\begin{align}
    p(y_t | \{x_t\}_{t=1}^M, \{y_t\}_1^{t-1}) = g(y_{t-1}, h^d_{t-1}, c_t)
\end{align}
\end{linenomath}
where $c_t$ (often named the context) is a result of the query procedure. Context vector can be seen as a continuous bag of weighted features of encoder hidden states $\textbf{h}^e$ \citep{chan2016listen}.

Finding the appropriate query procedure is a area of current research. In a simple form, the query is performed in the following steps. First, we compute a similarity measure between $q_t$ and each of $h^e_k$ for $k\in\{1, ..., M\}$, e.g. the kernel inner product $e_{t,k} = \langle q_t, h^e_k \rangle_{W^a} = q_t' W^a h^e_k$ where $'$ indicates transpose and $W^a$ is a (square) matrix. We normalize the similarity over $k$ by
\begin{linenomath}
\begin{align}
    w_{t, k} = \frac{\exp(e_{t,k})}{\sum_j \exp(e_{t,j}) }.
\end{align}
\end{linenomath}
The context is computed as the weighted linear combination of the encoder hidden states
\begin{linenomath}
\begin{align}
    c_t = \sum_{k=1}^M w_{t,k} h^e_{k}.
\end{align}
\end{linenomath}

In the original paper that proposed attention, \citet{bahdanau2014neural} commented that the approach of taking a weighted sum of all the annotations (e.g. $\{h^e_k\}_{k=1}^M$) is as if computing an expected annotation, i.e. $\E[h^e]$, over all the possible alignments. Here, alignment is defined by the weights $\{w_{t,k}\}_{k=1}^M$, which essentially aligns the $t$ decoder with the entire encoder annotation sequence $\{h^e_k\}_{k=1}^M$. In this sense, $w_{t,k}$ is a estimated probability that the target word $y_t$ is aligned to or translated from a source work $x_k$. Thus the $t$-th context vector $c_t$ is the expected annotation over all the annotations with probability $w_{t,k}$.

Two kinds of attention models have been proposed. The hard attention model uses the weights to randomly select one location \citep{mnih2014recurrent, ba2014multiple}; the soft attention model uses the weights to form a convex combination of many locations \citep{bahdanau2014neural, ba2014multiple}. In this work, we choose not to use stochastic ``hard" attention for two reasons. First, stochastic attention requires sampling one input stream to make a decision. If both the visual and audio input are noisy, the stochastic attention model has to choose one that is more informative in making a decision. However, if the audio noise and visual noise are independent, the information from each stream may compensate the other. Therefore, we choose to use a deterministic ``soft" attention model which dynamically combines multiple streams of data. Second, stochastic attention model involves a intractable objective function with a multinomial latent variable. Monte Carlo and REINFORCE \citep{williams1992simple} are two common methods for solving this problem. However, it is well known that the both estimation procedures are slow and unstable. On the contrary, soft attention model is fully differentiable and can be optimized using stochastic gradient descent.

\subsubsection{Spatial-Temporal Attention}
%In sequential prediction problems, sequence-to-sequence model is a ground-breaking work \cite{cho2014properties, cho2014learning}. However, RNN such as LSTM still suffers from difficulty in memorizing long-range relations, especially when a lot of information is cluttered in a sequence. The attention mechanism solved this problem by providing a skip-connection and let the decoder focus on a sub-region of the input \cite{bahdanau2014neural, chorowski2015attention, mnih2014recurrent, ba2014multiple}. Instead of memorizing the entire information of the past sequence, attention filter let the RNN remember the location of the past information. This greatly reduce the information load of the RNN.

% \cite{donahue2015long} apply LSTMs to generate video descriptions. It represents images as a single feature vector from the top layer of a pre-trained convolutional network. 

The temporal attention mechanism enables RNN to selectively focus on a subset of the input sequence. In multimodal learning, in addition to temporal attention, we aim to selectively focus on one (or a few) modalities of current time step and the past, which requires a spatial-temporal attention mechanism. 

A relevant problem is image caption generation whose task is to transcribe an image into a word sentence that best describes the image. In \cite{xu2015show}, a context vector is used to dynamically select relevant parts of a image to generate a word at time $t$. The context vector is a function of the RNN hidden states of time $t-1$ and the image features. \cite{xu2015show} define a mechanism that generates a positive weight for each image feature location, which can be interpreted as the probability that location is the right place to focus for producing the next word in a sequence. 

The problem we address here is different in a few places. First of all, in  \cite{xu2015show} the image input to the decoder RNN is the same for all time steps. The attention model gives time dependent weight to each location of the image based on the decoder hidden state input and decoder model output of $t-1$. We try to address a problem where at each time step the inputs (e.g. audio, motion, image) is different. This makes our model applicable to, e.g. online video description. Secondly, in  \cite{xu2015show} the attention is imposed on the static extracted image features. In our case, because at each time step the inputs are different, we defer attention to each local expert's decision. Third, although the image features are at different locations, they are same type of data and carry similar information. In our model, we have input features of totally different data types, e.g. image, motion and audio. Last, the image features are spatially dependent and the dependency does not change. In video, the spatial dependency of different input streams are dynamic and asynchronous. In summary, we use a dynamic-input cross-modality attention.

Another relevant work is dual-stage attention RNN (DA-RNN) \citep{qin2017dual}. Dual-stage attention solves two unique problems different from classical attentions. First, an input attention is used to attentively extract relevant information from multiple parallel exogenous input sequences instead of one input sequence. Then a temporal attention mechanism is used to select relevant encoder hidden states generated by input attention across all time steps. The decoder outputs a time-series prediction instead of a classification. Experiments show that DA-RNN is effective in time-series prediction and robust to noisy inputs. 

RNN performance will deteriorate rapidly as the length of input sequence increases, \citet{qin2017dual} used a temporal attention in the decoder to adaptively select relevant encoder hidden states across all time steps. They use the concatenate score function in \citep{luong2015effective}. %Rarely have I seen this score function been used instead of the generalized inner product. \lijiang{Could this be the key to successful time-series prediction?}
\begin{linenomath}
\begin{align}
    l_t^i = \mathbf{v}_d^T \text{tanh}(\mathbf{W}_d[\mathbf{d}_{t-1};\mathbf{s}_{t-1}'] + \mathbf{U}_d \mathbf{h}+i), \quad 1\leq i \leq T
\end{align}
\end{linenomath}
and 
\begin{linenomath}
\begin{align}
    \beta_t^i = \frac{\exp(l_t^i)}{\sum_{j=1}^T \exp(l_t^j)},
\end{align}
\end{linenomath}

In the encoder, $\x_t \in \R^n$ is the observation of $n$ exogenous driving input sequences at time $t$. In a normal RNN, the input sequences $\{\x_t\}_{t=1}^T$ is fed to a hidden unit such as LSTM which is updated by
\[
    \h_t = f(\h_{t-1}, \x_t).
\]
\cite{qin2017dual} proposed to use $\h_{t-1}$ and $\x_t$ to generate a set of $n$ weights for $x_t$:
\begin{linenomath}
\begin{align}
    e_t^k = \mathbf{v}_e^T \text{tanh}(\mathbf{W}_e[\h_{t-1};\mathbf{s}_{t-1}]+\mathbf{U}_e \x^k)
\end{align}
\end{linenomath}
and
\begin{linenomath}
\begin{align}
    \alpha_t^k = \frac{\exp(e_t^k)}{\sum_{i=1}^n \exp(e_t^i)}
\end{align}
\end{linenomath}
where $\mathbf{v}_e \in \R^T, \mathbf{W}_e \in \R^{T \times 2m}$ and $\mathbf{U}_e \in \R^{T\times T}$ are parameters to learn. Using the weights $\alpha_t$, a new input is
\begin{linenomath}
\begin{align}
    \tilde{\x}_t = \x_t \odot \alpha_t.
\end{align}
\end{linenomath}
Then the hidden state at time $t$ is updated as
\begin{linenomath}
\begin{align}
    \h_t = f(\h_{t-1}, \tilde{\x}_t).
\end{align}
\end{linenomath}
With the proposed input attention mechanism, the encoder can selectively focus on certain driving series rather than treating all the input driving series equally.

Note in \citep{qin2017dual} the prediction is on $\y_t \in \R$. But in denoising, the prediction is on $\y_t \in \R^D$ where $D$ is the number of frequency bins.

% In the encoder, we introduce a novel input attention mechanism that can adaptively select the relevant driving series. In the decoder, a temporal attention mechanism is used to automatically select relevant encoder hidden states across all time steps. For the objective, a square loss is used. With these two attention mechanisms, the DA-RNN can adaptively select the most relevant input features and capture the long-term temporal dependencies of a time series.

% In a recent work, I built a multimodal speech activity detection system using spatial-temporal attention. My idea is very similar to the dual-stage attention RNN model.

\subsubsection{Attention mechanism in audio-visual problems}

Temporal attention mechanism has been used to improve RNN performance for speech related problems. \cite{chan2016listen} proposed Listen, Attend and Spell (LAS), which is an audio-only speech recognition model based on sequence-to-sequence learning framework with attention. The model learns to transcribe an audio sequence signal to a word sequence, one character at a time. It consists of an encoder RNN, and a decoder RNN. The encoder RNN is a pyramidal RNN which converts low level speech signals into higher level features. The decoder is an RNN that converts high level features into output utterances by specifying a probability distribution over sequences of characters using the attention mechanism. The encoder and decoder are trained jointly. The key contribution of \cite{chan2016listen} is the pyramidal RNN model for the encoder, which reduces the number of time steps that the attention model has to extract relevant information from. 

%\cite{assael2016lipnet} proposed a sentence level video-only speech recognition model. They combine CNN and LSTM and Connectionist Temporal Classification to classify word labels. 

% In \cite{chan2016listen}, the pyramidal RNN structure is key to training long sequences. \cite{chan2016listen} use pyramial structure and scheduled sampling training to make the LAS perform well on testing set, achieves $14.1\%$ WER on a voice search task, without a dictionary or a language model.

RNN has been well known for its power in modeling sequential relation. It is also hard to train and can easily overfit. The attention mechanism has been used as a way to reduce overfitting. \cite{chan2016listen} commented that without the attention mechanism, the model overfits the training data significantly and memorizes the training transcripts without paying attention to the acoustics.

\cite{chorowski2015attention} proposed an attention based audio-video speech recognition model. Their model has a two-stream network with temporal attention in each stream. Each stream's encoder consists of a feature extractor, and a RNN encoder. The RNN encoder takes per-frame input from feature extractor in reverse time order. In the end, the two streams' RNN sequence outputs and final state are fed into a RNN decoder to generate texts. They found the attention mechanism is critical for the speech recognition system to work. Without attention, the model appears to forget the input signal, and produce output sequence that correlates very little to the input. \cite{chorowski2015attention} use separate temporal attention to each modality in the decoder. However, the decoder gives equal  weights to each modality when generating outputs. That is, each modality receives a sum of attention equals $1$. Suppose one modality is corrupted, the weight on that modality is not adjusted. In this regard, \cite{chorowski2015attention} does not consider the variation of information in modalities.

% \cite{chung2016lip2} claims that the dual-attention mechanism allows the model to extract information from both audio and video inputs, even when one stream is absent, or the two streams are not time- aligned. The benefits are clear in the experiments with noisy or no audio (Section 5).

%We use a attention expert to explicit evaluate each input stream, and assign weights to that stream.

% [someone] proposed a algorithm for video description. The model looks at the entire video at every step to generate a new word. [someone] proposed a multimodal video description algorithm using attention. However, both the temporal and spatial attention have to look at the entire video at each step to generate a new word. 

%\lijiang{Add discussion about \citet{hori2017attention}. To fuse multimodal information, \citet{hori2017attention} proposed method extends the attention mechanism. The approach can pay attention to specific modalities of input based on the current state of the decoder to predict the word sequence in video description.}

Here we present a spatial attention based sensor fusion model as a principled way of combining different modality of signals (i.e. audio and video) for e.g. speech enhancement. The model is trained end-to-end, and simultaneously learns spatiotemporal audio-visual features and a sequence model.

\subsection{Mixture of  Sub-task Experts}
Dynamically combining multiple functions in supervised learning can be traced back to \emph{mixture of experts} model \citep{jacobs1991adaptive}. A mixture of experts model is driven by the assumption that a set of training cases may be naturally divided into subsets that correspond to distinct tasks. However, the interference between different subsets of tasks would lead to slow learning and poor generalization. Such interference can be reduced by using a system composed of several different \emph{expert functions} where each one is trained for a subset of tasks. A \emph{gating function} is trained to decide which of the experts should be used for each training case. Instead of a hard decision such as decision tree, the gating function makes probablistics decision by assigning mixing weights to the experts. Neural networks can be used for both the expert functions and gating function, which is known as \emph{mixture density network} (MDN) \citep{bishop1994mixture}. % mixture density network is cited in MLAPP.
One advantage of MDN is it can approximates a flexible family of distributions, including distribution of multiple modes.

\begin{figure}[t!]
    \centering
    \includegraphics[width=0.6\textwidth]{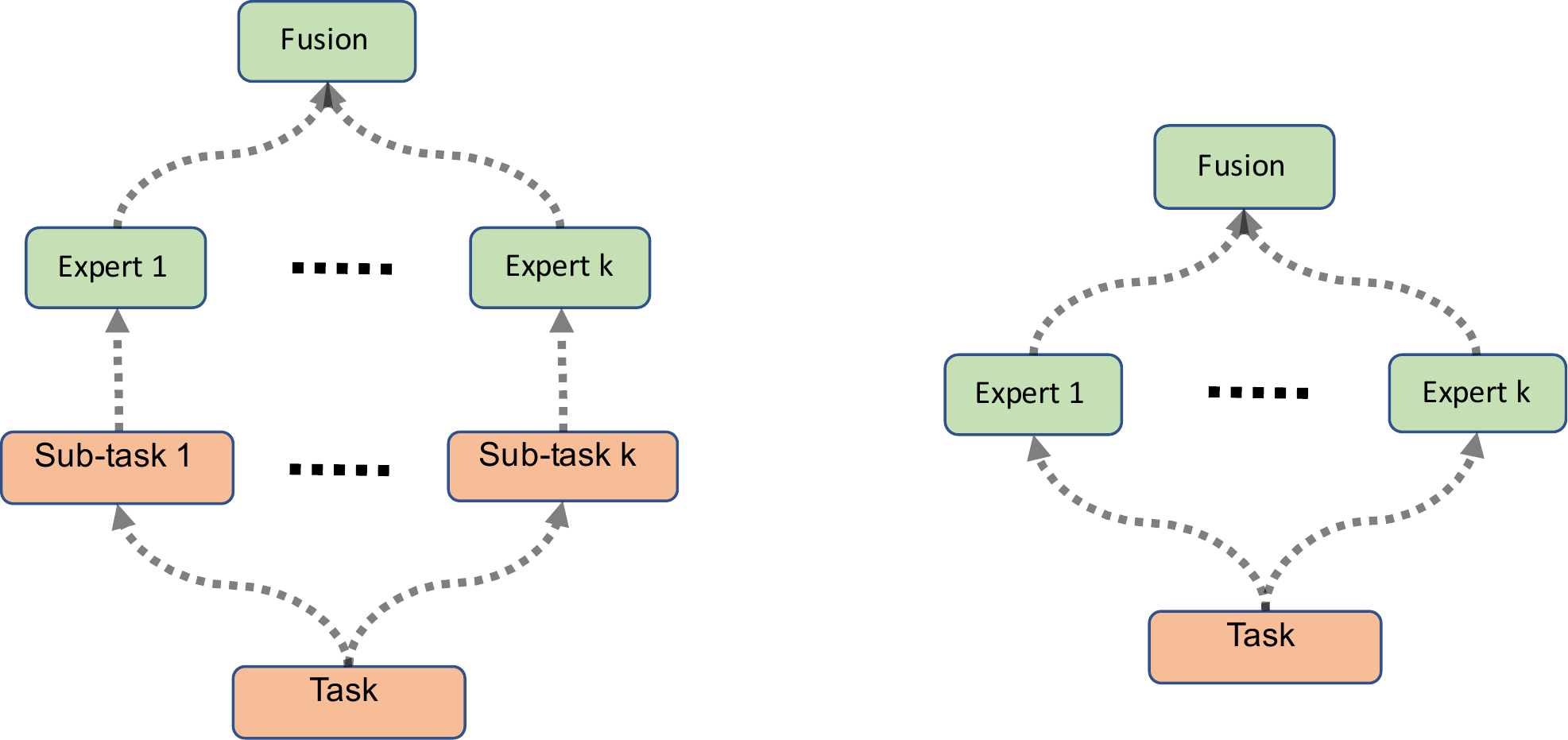}
    \caption{Mixture of sub-task experts (left) and mixture of experts (right). In mixture of experts, each expert receives the identical task. In mixture of sub-task expert, each expert receives a distinct subset of the task.}
    \label{fig:sub-tasks}
\end{figure}

\cite{jacobs1991adaptive} want a system to learn how to allocate cases to experts. They designed a loss function such that the gating network allocates a new case to one or a few experts, and if the output is incorrect, the weight changes are localized to these experts and the gating network. The experts are therefore local in the sense that the weights in one expert are decoupled from the weights in other experts. In addition they will often be local in the sense that each expert will be allocated to only a small local region of the space of possible input vectors.

Different from dividing all training cases into subsets of tasks, we observe that a training case can be divided into \emph{parallel sub-tasks} (\autoref{fig:sub-tasks}). For example, in using two hands to open a jar, the parallel sub-tasks are left-hand movement and right-hand movement. In speech, the lip movement and audio are data of two sub-tasks, where speech is the super-task.

Many recent researches have implicitly utilized such a parallel sub-task design. In activity recognition, the two-stream network \citep{simonyan2014two} has a sub-task design for extracting motion and image features. The shared decoder works as a fusion function which takes concatenated inputs from sub-networks. In audio-visual speech recognition, \cite{chorowski2015attention} use two encoder neural networks for processing audio and video signals. The two also share one decoder which takes concatenated inputs from the two encoders. %\cite{kim2017collaborative} use modular neural network design to combine various speech enhancement modules in parallel. Each module is a deep neural network specialized on a particular speech enhancement job. A autoencoder works as a gating network to select the best enhancement module for each input example.
In these works, the gating function is implicit in the decoder. Such a blackbox methods does not explicitly assign sub-tasks to experts, and the error backpropogation may cause global changes to expert functions.

We make three extensions to MDN for multimodal sensor fusion. First, we separate expert networks into different functional components by dividing a training case into parallel sub-tasks, one for each sensor. Sub-tasks could have same objective or partially overlapping objectives. Each expert network receives input from only one unique sensor, hence is forced to solve only one sub-task. The gating network receives all sensor inputs to decide a soft mixing weight. This contrasts a boosting algorithm, e.g. mixture of expert, where each expert receives the complete and identical task.

Second, in the mixture of density network, one function (i.e. neural network) is used to predict the parameters of all of the component densities as well as the mixing coefficients, so the hidden units between input and output layers are shared among the input-dependent functions. We further divide the functions into different modular neural networks, such that each expert learns a unique function, e.g. neural network for the assigned sub-task.

A major concern of mixture of experts model is co-adaption of experts. The third contribution of our work is to re-address this issue by promoting multiple experts to learn similar features, and also learning dissimilar features at the same time; we propose a mechanism to partition latent variables into co-adaption set and independent set; see \autoref{sec:co-learning}.

% Because each modality's input signal is naturally different from the others, e.g. audio vs image, using a common model to process all the input is not going to effectively extract the useful features of a modality. Therefore, we consider a modular neural network design for processing multimodal inputs.

\subsection{Recurrent Attention Filter for Multimodal Sensor Fusion} \label{sec-recurrent-attention-multimodal-sensor-fusion}
In this section we introduct recurrent attention filter for sensor fusion. We try to marry the strength of spatial-temporal attention and mixture of sub-task experts. For demonstration, we use speech activity detection where the inputs are audio, image, and motion, and outputs are binary labels. This model can be converted to a, e.g. speech and video enhancement system where outputs are, e.g. spectrograms and images.

Let $\{\x_1, ..., \x_T\}$ be a sequence of input signals. Partition $\x_t$ into $\{\x_t^{(1)}, \x_t^{(2)}, \x_t^{(3)}\}$ which are audio, image, and motion features, respectively. These features are functions, e.g. neural networks, of input signals. Let $\y = \{y_1, ..., y_T\}$ denote the sequence of speech activity labels, i.e. $\{0, 1\}$. Let $\x^{(m)}_{\leq t}$ and $\y_{\leq t}$ denote the subsequences $\{\x^{(m)}_1, ..., \x^{(m)}_t\}$ and $\{y_1,...,y_t\}$. We model $y_t$ as a Bernoulli random variable with conditional distribution:
\begin{equation}
    \P(y_t = 1|\x_{\leq t}) = \langle g(\x_{\leq t}), [f_1(\x^{(1)}_{\leq t}), f_2(\x^{(2)}_{\leq t}), f_3(\x^{(3)}_{\leq t})] \rangle,
\end{equation}
where $\langle \cdot , \cdot \rangle$ is Eculidean inner product. Each \emph{expert} function $f_m(\cdot)$ is a neural network which predicts the conditional distribution of $y_t$ given input from modality $m$ only:
\begin{equation}
    f_m(\x^{(m)}_{\leq t}) = \P(y_t = 1|\x^{(m)}_{\leq t}).
\end{equation}
The \emph{spatial attention} function $g: \mathbb{R}^n \mapsto \mathbb{R}^m$ is also a neural network which specifies a mixing weights for each $f_m$. Input to $g$ are features of all modalities, as $g$ has to output weights on all modalities. The output weights are non-negative and sum to 1. Therefore, $y_t | \x_{\leq t}$ has a mixture of Bernoulli distributions, where each component is a conditional distribution:
\begin{align}\label{eq:WAL_objective}
p(y_t | \x_{\leq t}) = \sum_m p(m|\x_{\leq t})p(y_t|\x^{(m)}_{\leq t}).
\end{align}
%Notice $\{y_t\}$ are not independent and not identically distributed due to temporal dependency within $\x_t$. However, the sequential dependency between $\x$ provides additional information for modeling and prediction.
In the \emph{recurrent attention mixture model} (\autoref{eq:WAL_objective}), for each expert function we apply temporal attention on the input $\x^{(m)}_{<t}$, such that $f_m(\x^{(m)}_{\leq t})$ can look back in time and focus on relevant time segments. The recurrent attention mixture model can be solved using stochastic gradient descent. RNN with attention is often slow to train and has high memory consumption. Therefore the RNN with temporal attention can be replace with 1-D dilated convolutional neural network.

\begin{figure}[!t]
\begin{center}
% \fbox{\rule{0pt}{2in} \rule{0.9\linewidth}{0pt}}
   \includegraphics[width=0.5\linewidth]{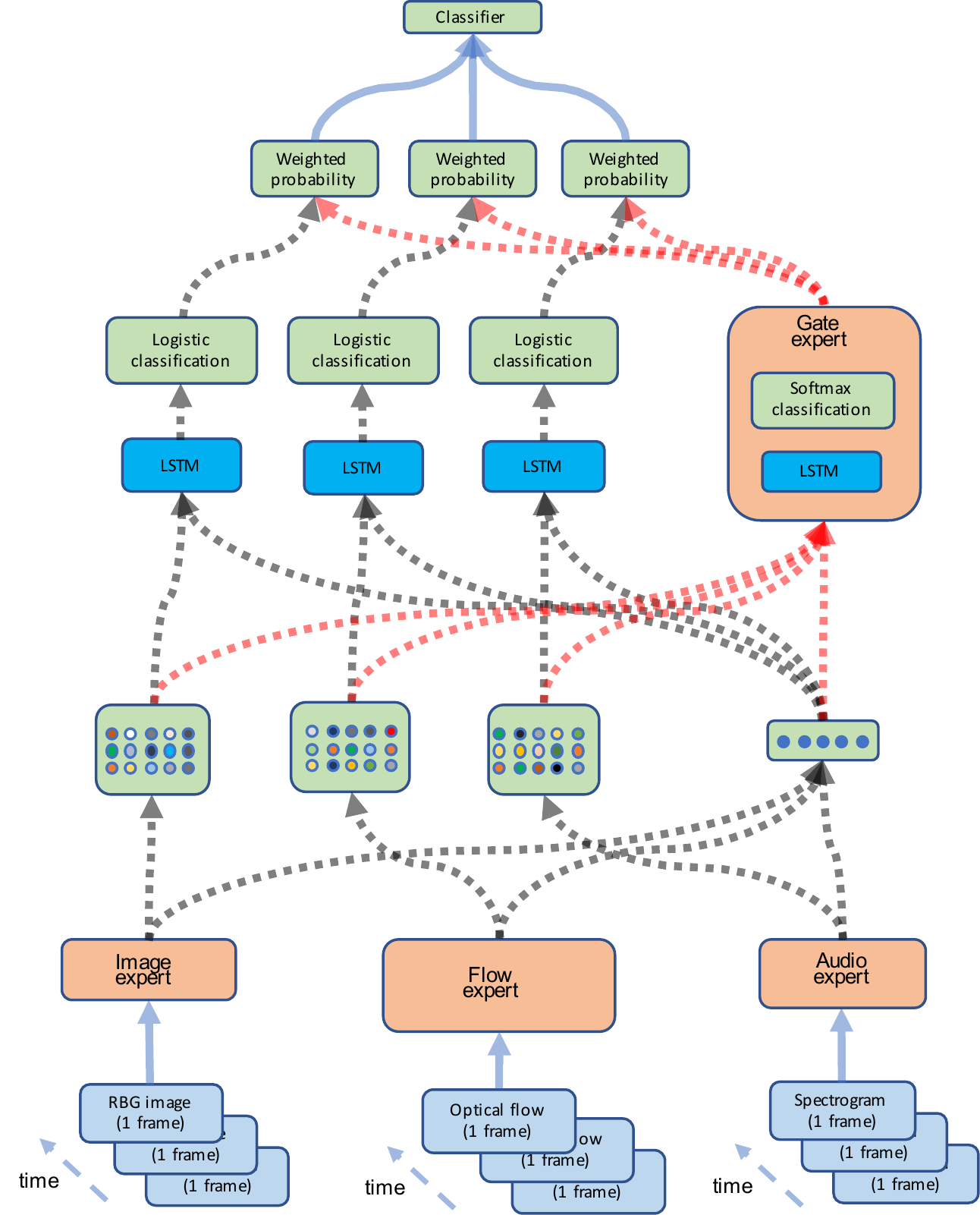}
\end{center}
   \caption{Regularized recurrent attention multimodal sensor fusion model.}
% \label{fig:long}
\label{fig:regularized-recurrent-attention-filter}
\end{figure}

A simplified version is akin to Markov mixture of expert \citep{meila1996learning} where current prediction is independent of the past given a latent random variable:
\begin{align}\label{eq:Markov_attention}
p(y_t | \x_{\leq t}) = \sum_m p(m|\x_{t}, \h'_{t-1})p(y_t|\x^{(m)}_{t}, \h^{(m)}_{t-1})
\end{align}
where $\h'$ and $\h^{(m)}$ are RNN hidden units in attention and expert networks. The \emph{Markov attention mixture model} (\autoref{eq:Markov_attention}) is shown in \autoref{fig:regularized-recurrent-attention-filter}. Regularization is discussed in \autoref{sec:co-learning}. 

We can further simplify the model by assuming $\y_t \perp \x_{< t} | \x_t$ such that 
\begin{equation}\label{eq:iid_WAL_objective}
p(y_t | \x_{\leq t}) = \sum_m p(m|\x_{t})p(y_t|\x^{(m)}_{t}).
%    \P(y_t = 1|\x_{\leq t}) = \P(y_t = 1|\x_{t}) = \langle g(\x_t), [f_1(\x^{(1)}_{t}), f_2(\x^{(2)}_{t}), f_3(\x^{(3)}_{t})] \rangle.
\end{equation}
This simplified model (\autoref{eq:iid_WAL_objective}) is a \emph{conditional attention mixture model} --- each time step is a independently distributed mixture of conditional Bernoulli distributions. We can follow the standard EM receipt \citep{bishop2006pattern} to solve the Markov attention mixture model and conditional attention mixture model (see Appendix A). 

\subsubsection{Comments}
In multimodal sensor fusion, each expert function would process one of the parallel sensor input, rather than a subset of all the training cases as in mixture of experts (\autoref{fig:sub-tasks}). The gating function is replaced with a spatial attention function, which at each time step, generates soft attention over sensors. 

\citet{baltruvsaitis2018multimodal} suggested that late fusion ignores the low level interaction between the modalities. However, \citet{ngiam2011multimodal} found that it is difficult to capture cross-modality relation with low level features, because within-modality correlation is much stronger than between-modality correlation. They suggested that fusion at higher level would encourage learning cross-modality features, because higher level features may have less within-modality correlation than raw features. 

In our model, the gating network could take input features from any level of the expert networks. It is possible to apply a hybrid of late and early fusion by taking both high and low level features. %\lijiang{(Can add discussion on style transfer and perceptual loss.)} 
Hence our model can be considered as a model agnostic hybrid fusion. These features are used to create dynamic mixing weights, similar to a spatial attention mechanism.

The spatial attention function $g$ is similar to soft attention \citep{xu2015show}. But there are two key differences. First, $g(t)$ does not have input from $\x_{>t}$; this is critical because $g$ does not rely on future information to make current decision, which allows for online decision making. Second, for $t+1$, $g$ takes new input $\x_{t+1}$ in addition to previous inputs $\x_{\leq t}$, hence a filter.  %the input sequence is dynamic, and $g$ combines all sources' input features from beginning to current time.

The recurrent attention mixture model is similar to mixture density network \citep{bishop1994mixture}. The major difference is that we use dedicated neural networks to predict each component density (i.e. expert) and mixing coefficients (i.e. gate), thus maximizing the utility of each sensor's input. To recognize the covariance between different sensors, we design a co-learning mechanism as explained in next section. 

\subsection{Co-learning Latent Features}\label{sec:co-learning}
Each expert function $f_m$ is trained to maximize the utility of its sensor input. Although sensor inputs are different, they are likely to have some similar latent features. For example, when uttering some words, part of the lip movements (i.e. visemes) are coordinated with the sound (i.e. phoneme). In this sense, two input modalities and not independent. We can make an assumption that for the speech activity detection task, for each experts, there are some common features that ``represent" some states; for example, a fixed set of nodes in both experts' hidden layer are activated when speech is found. If we can apply some prior information which encourage latent variable sharing, we may extract more robust features for the classification task.

%In a speaker identification task, \citet{ren2016look} proposed a multimodal LSTM architecture which shares weights across time steps as well as across modalities

To this end, we consider two techniques from statistical learning. One is to apply a penalty term to part of hidden unit outputs to encourage the output of different experts be similar to each other. For example, we can add a term to the loss function which is proportional to the sum of squared distances between outputs and average of outputs. Alternatively, instead of tying hidden unit outputs we can put penalty on weights to encourage some weights are tied between different expert networks. 

A second approach is to define a structure prior of the generative model via probabilistic graphical model. We will discuss this approach extensively in \autoref{sec-msrnn}.

\subsubsection{Distance based regularization}
For each expert network, choose one hidden layer. Let $\z_1,...,\z_M$ denote the hidden unit outputs of these layers of the $M$ experts. Let $\z_m^*$ denote the first $n$ hidden nodes of $\z_m$. We assume that
\begin{align}
    \E[\z_m^*] = \zeta.
\end{align}
That is, for all input sensors, $\z^*_m$ has the same expectation. Thus 
\begin{align}
    \hat{\zeta} = \frac{1}{K}\sum_{k=1}^K z_k^*
\end{align}
is an unbiased estimator of $\zeta$, where $z_k^*$ is a sample of $\z_k^*$.
Define the co-learning loss as
\begin{align}
    \mathcal{L}_{co} = \sum_{m=1}^M \lambda_m\|z_m^* - \hat{\zeta}\|_2^2.
\end{align}
$\mathcal{L}_{co}$ is a sum of squared $L2$ norms; $\lambda_k$ is a sensor specific penalty parameter selected by cross-validation. Adding $\mathcal{L}_{co}$ to the loss function has the effect of shrinking $\z_k^*$ towards $\zeta$, as in Tikhonov regularization, to prevent overfitting and stabilize parameter estimation.

We want to point out that if we assume $\z^*$ are Gaussian random variables, then $\mathcal{L}_{co}$ is equivalent to impose a Gaussian prior distribution over $\zeta$. In this sense, the $\mathcal{L}_{co}$ estimate of $\z^*$ is a MAP estimate. The advantage of formulating $\mathcal{L}_{co}$ as Tikhonov regularization is that we can replace $\hat{\zeta}$ with a moving average of the current mini-batch of samples in stochastic gradient descent.

\subsection{Recurrent Attention Filter for Audio-Visual Speech Separation}
It is easy to modify our model from speech activity detection to speech separation. For speech separation, a common technique is to use a ideal ratio mask (or a ideal binary mask), which is a element-wise ratio (or binarized ratio) between clean and noisy spectrogram. This mask is then multiplied with the input spectrogram to get a denoised spectrogram. The audio input could be either complex spectrogram or magnitude spectrogram. In the case of complex spectrogram, two masks will be generated for real and imaginary component respectively. When using magnitude spectrogram, one mask will be generated, and the phase of noisy input will be used to as denoised phase. The speech activity detection model discussed above can be seamlessly transformed into a speech separation model.

\section{Model Based Sensor Fusion: Separating Modality Invariant and Modality Dependent Information}\label{sec-msrnn}
In the previous section, we discussed a deep learning model for multimodal sensor fusion. Towards the end, we propose a co-learning idea which encourages co-adaption and independent learning of each sensor at the same time. In the following sections, we propose probablistic models which combines multiple sensors' signals into separate modality-dependent and modality-invariant features. We begin by introduce a variational RNN (VRNN) model where the transition between hidden units are stochastic. The VRNN model is a non-Markovian model in the sense the conditional independence assumption is broken in both transition and emission. We discuss this property by comparing VRNN with hidden Markov model and linear dynamic system. Using VRNN, we build a multimodal VRNN which imposes a structural prior on the generative model. As a result, the modality dependent and modality invariant factors are encouraged to separate into different latent variables. We emphasis that as oppose to PCA or VAE where latent factors are not known a priori without looking at the posterior distribution, our model explicitly matches latent variables with concepts. This is a result of the explicit graphical model structure.

The non-Markovian model brings a price. One direct outcome is that the latent state no longer contains \emph{all the information} for generation or transition. In this sense, it is not a state-space model in the classical sense. As the latent state does not contain all the information, it is questionable to use the latent state for many downstream tasks; for example, control and planning. Another outcome is that the time derivative information is not captured by the VRNN model as the latent variable only focuses on recovery of the current observation. %(\lijiang{this statement is problematic, needs more clarification; see \citep{karl2016deep}.}) 
%Therefore, we propose a multimodal deep state-space model to solve this issue.

\subsection{Motivation}
Our motivation is driven by a few observations. First, we observe from human representation learning that feature representations are either modality-invariant or modality-dependent: when you listen to a person, your ears hear the speech, and your eyes watch the person's face, then you jointly use the visual and audio signals to decode the content of the speech which is embedded in both audio and visual signals, and other information such as speaker's visual identity and vocal accent which are uniquely embedded in visual or audio signal. With multiple input channels, we can infer the speech content with more precision than with a single input. For this reason, if we consider audio and visual signals are generated by some latent explanatory factors, it is natural to partition the latent factors into modality-dependent and shared subsets. 
%\lijiang{LDS and RNN are naturally suitable for temporal data.} 

A second observation is many signals have a temporal structure. If we naively separate each time step as a independent unit for analysis, the data does not carry the same information. A typical approach is to formulate a sequence of latent random variables with Markov property to generate the observed sequence. Two common models are hidden Markov model and linear Gaussian models. However, neither of which are well-suited to modeling long-term dependencies and complex probability distributions over high-dimensional sequences. Neural network models such as RNN have seen many successes in modeling complex sequences with long dependency. One limitation with RNN is structural variations are captured by deterministic transition.
%As discussed in \citep{chung2015recurrent}, when there is strong structural variation (i.e. high signal-to-noise ratio), it is mixed with random noise in inputs and outputs. 
While a RNN can increase its memory capacity by increasing the size of its hidden unit, it is more likely to overfit to training data. There is recent evidence that when complex sequences such as speech and music are modeled, the performances of RNNs can be dramatically improved when uncertainty is included in their hidden state. In this work, we combine stochastic latent variable with RNN to take advantage of both methods in a coherent way. 

Considering learning a generative model for video of speech. Learning interpretable representations for such data, and comparing them as the speech content or the speaker identity are changed, gives useful high-level tools for speech recognition and speech synthesis. Even though each image is encoded by thousands of pixels and each audio segment is represented as hundreds of frequency bins, the data lie near a low-dimensional nonlinear manifold. A useful model must not only learn this manifold but also provide an intepretable representation of the speech dynamics. A natural representation from speech is that the human speech audio is divided into \emph{phonemes}, and speech video is divided into \emph{visemes} \citep{bear2017phoneme}. Therefore an appropriate model might switch between discrete states with each state representing the dynamics of a particular action. These two learning tasks, identifying an speech manifold and a structured dynamics model, are complementary. We want to learn the speech manifold in terms of coordinates in which the structured dynamics fit well.

 \begin{figure}[t!]
     \centering
     \begin{subfigure}[b]{0.4\textwidth}
         \includegraphics[width=\textwidth]{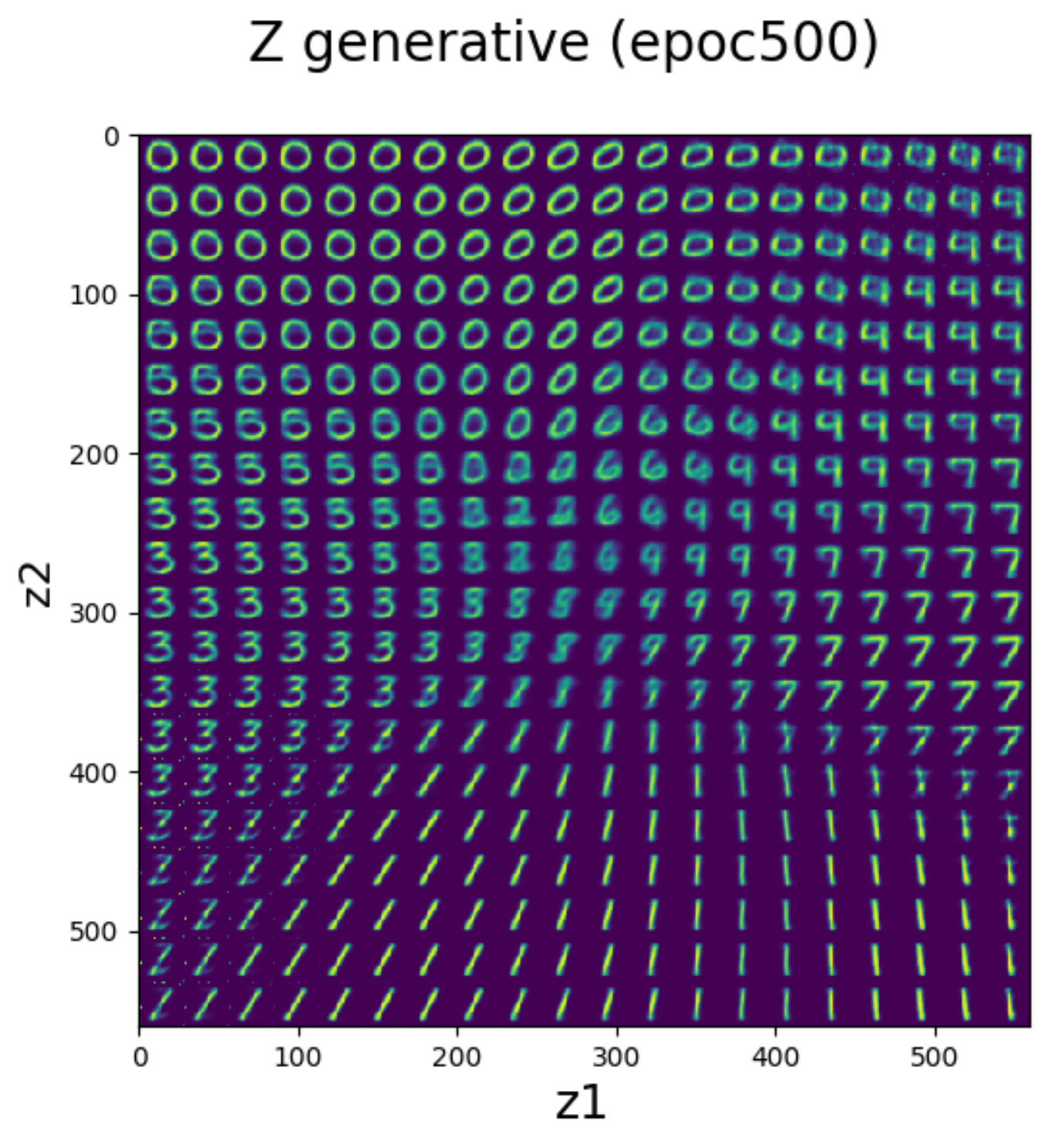}
         \caption{Latent space of VAE for MNIST digits.}
         \label{fig:vae-mnist-latent-space}
     \end{subfigure}
     \qquad
     %~ %add desired spacing between images, e. g. ~, \quad, \qquad, \hfill etc. 
       %(or a blank line to force the subfigure onto a new line)
     \begin{subfigure}[b]{0.43 \textwidth}
         \includegraphics[width=\textwidth]{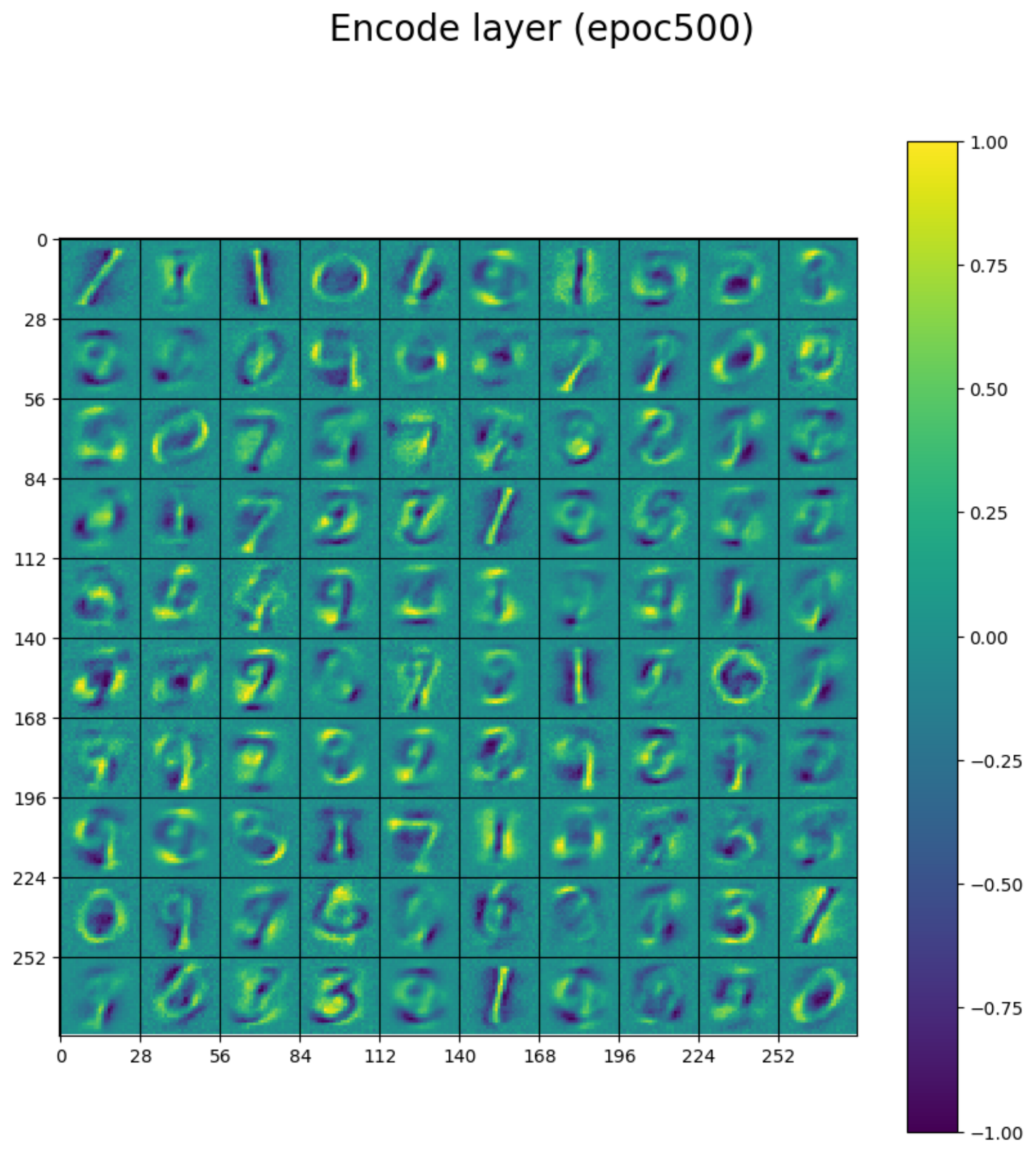}
         \caption{Features for MNIST digits.}
         \label{fig:vae-mnist-latent-feature}
     \end{subfigure}
     \caption{Separating content and style from MNIST digits.}
     \label{vae-mnist}
 \end{figure}

In our preliminary experiments, we used variational autoencoder (VAE) to map images of digit according to content and style by learning from the MNIST dataset. In \autoref{fig:vae-mnist-latent-space} we project the latent space to a 2 dimension space, and we can see clear clustering of digits, i.e. contents. The row dimension and column dimension show the variation of style of each digit. We also plot the hidden layer activation map in \autoref{fig:vae-mnist-latent-feature}, and we can observe meaningful patterns corresponding to digits.

\subsection{Related Works}
\subsubsection{Inference in Graphical Models with Conjugate Prior}
In Gaussian linear dynamic system model with linear Gaussian observation, because the observation model $p(y|x)$ is conjugate to the latent variable model $p(x)$, e.g. Gaussian, then the optimal approximate distribution is naturally Gaussian, and we can use efficient message passing algorithms to perform exact inference. However, when the observation model is not conjugate to the latent variable model, these algorithmic structure break down. A solution is to use general variational inference such as mean-field method. Mean-filed method makes strong independence assumption on the latent distribution for ease of computation, which lead to greater gap between the marginal likelihood and variational lower bound. 

%\lijiang{Amortized variational inference and recognition networks.}

\subsubsection{Variational Autoencoder}
%\lijiang{Discuss VAE principle.} 
Two common approaches for solving latent variable models are variational inference and Markov Chain Monte Carlo methods. Recently, stochastic variational inference \citep{hoffman2013stochastic} has been successfully combined with deep neural networks into a class of deep Gaussian models named Variational Autoencoder \citep{kingma2013auto, rezende2014stochastic}. Different from a vanilla autoencoder, a VAE introduces a set of latent random variables $\z$ to explain the variations in the observed variables $\x$. Their joint distribution is defined as:
\begin{linenomath}
\begin{align}
	p(\x,\z) = p(\x|\z)p(\z).
\end{align}
\end{linenomath}
The VAE typically parameterizes $p(\x|\z)$ with a highly flexible function approximator such as a neural network. The neural network allows for highly non-linear mapping from $\z$ to $\x$ which is a powerful and unique feature of VAE. However, introducing a highly non-linear mapping from $\z$ to $\x$ results in intractable inference of the posterior $p(\z|\x)$. VAE uses a variational approach to approximate the posterior distribution $p(\z|\x)$ while incrementally raises the evident lower bound:
\begin{linenomath}
\begin{align}
	\log p(\x) \geq \E_{q(\z|\x)} \log p(\x|\z) - KL(q(\z|\x) | p(\z)),
\end{align}
\end{linenomath}
where $q(\z|\x)$ is a distribution modeled using a inference neural network.

The generative model $p(\x|\z)$ and the inference model $p(\z|\x)$ are jointed optimized by maximizing the evidence lower bound. The expectation with respect to $q(\z|\x)$ is approximated stochastically in VAE. 

In VAE we estimate the distribution of latent variables $\z$ and use this information to reconstruct original signal $\x$. Because $\z$ is random rather than deterministic, VAE allows for reconstructing different $\x$ from different modes of $\z$ rather than a single point estimate as in regular auto-encoder. 

% \lijiang{Add more details on stochastic variational inference.}

\subsubsection{Recurrent Neural Network}
An RNN is a autoregressive model which takes a sequence input $\{\x_t\}_{t=1}^T$ to predict a sequence output $\{\y_t\}_{t=1}^T$. At each time step $t$, the RNN reads the input $\x_t$ and updates its hidden state $\h_t$ by
\begin{linenomath}
\begin{align}
	\h_t = f(\x_t, \h_{t-1}; \theta),
\end{align}
\end{linenomath}
where $f$ is a deterministic non-linear function with parameters $\theta$. Common choices of function $f$ can are gated activation functions such as LSTM or GRU. 

While the function $f$ is deterministic, we can add randomness to RNN by letting the output be a distribution. For example, assuming a sequence of lag $1$ autoregressive relation, i.e. $\x_t$ depends on $\{\x_{k}\}_{k=1}^{t-1}$. Conceptually, RNN models this sequence by parameterizing a factorization of the joint sequence probability distribution as a product of conditional probabilities such that 
\begin{linenomath}
\begin{align}
	p(\x_1, ..., \x_T) &= \prod_{t=1}^T p(\x_t | \x_{<t})\\
	&= \prod_{t=1}^T g(\h_{t-1}; \tau)
\end{align}
\end{linenomath}
where $g$ is a function which maps the hidden state $\h_{t-1}$ to the conditional probability distribution $p(\x_t | \{\x_{k}\}_{k=1}^{t-1})$.

We can model the output function $g$ as being composed of two parts. The first part $\varphi_\tau$ is a function that returns the parameter set $\phi_t$ given the hidden state $\h_{t-1}$
\begin{linenomath}
\begin{align}
	\phi_t = \varphi_\tau(\h_{t-1}).
\end{align}
\end{linenomath}
The second part of $g$ returns the conditional distribution of $\x_t$ as $p(\x_t| \{\x_k\}_{k=1}^{t-1}) = p(\x_t|\phi_t)$.

%Gaussian mixture model (GMM)  is a common choice for modeling a high-dimensional and real-valued distribution, especially for modeling structured output density model. For GMM, $\varphi_\tau$ returns a set of parameters including the mixture coefficients and the means and covariance matrices of each Gaussian components.

However, given $f$ is a deterministic function, a RNN model does not have the stochastic transition in a HMM. %\citet{chung2015recurrent} suggested that when there is strong structural variation in the input signal, RNN must be capable of modeling both small variations due to noise and also large variations due to signal. 
The lack of a mechanism to model the structural variation imposes a restriction on the RNN, as when it attempts to encode sufficient input variability to capture the signal variations, it inevitably overfits noise variations. In order to prevent overfitting, we must limit the capacity of the RNN and find another mechanism to model signal variations.

\subsubsection{Stochastic RNN}
Modeling sequential data is a domain of interest to representation learning. In a temporal-aligned RNN, given a deterministic transition function, the only source of variability is the output distribution $p(y_t|x_{\leq t}) = p(y_t | x_t, h_{t-1})$. On the other hand, a limitation of standard HMM is that it is poor at capturing long-range correlation between the observed variables \citep{bishop2006pattern}. Recently there is a trend in combining probabilistic models models such as state space model with deep neural networks \citep{chung2015recurrent}. 
%In a highly structured sequence of signals (e.g. speech), the structure generates large amount of variations. Thus the input contains two sources of variation from noise and signal. Because the random noise is independent from time, we can add temporal dependencies across timesteps to model the structural variation in signal. 
The key is to incorporate some stochastic hidden states to RNN. \citet{chung2015recurrent} introduced a sequence of latent random variables whose prior distribution at time step $t$ is dependent on all the preceding inputs via a RNN hidden state $\h_{t-1}$. A similar model is proposed by \citet{fraccaro2016sequential}. These are stochastic RNN models, not state-space models in the strict sense as they have broken the conditional independence assumption in the emission model:
\begin{linenomath}
\begin{align}
p(\x_{1:T}|\z_{q:T}) \neq \prod_{t=1}^T p(\x_t|\z_t).
\end{align}
\end{linenomath}

SRNN can be solved using variational inference algorithm. VAE at heart is a simple probablistic graphic model of joint distribution of two time independent random variables $\x$ and $\z$. When there is temporal relation, the assumption is broken. The recognition model would only try to encode the current observation into a latent variable and the generative model would decode the latent variable. Thus VAE would overlook the transition between latent states. In this sense, the VAE model overfits to the training data and discards the temporal information in the training data.

%\lijiang{We can train a VAE as a discriminative network with some prior constraints \cite{hsu2017learning, hsu2017unsupervised}. To do later.}

%\lijiang{We discuss deep nonlinear state-space model such as deep Kalman filter \citep{krishnan2015deep} and variational Bayes filter \citep{karl2016deep} in \autoref{sec:deep-state-space-model}. To do later.}

\subsubsection{State Space Model: HMM and LDS}
\begin{figure}[t]
\begin{center}
% \fbox{\rule{0pt}{2in} \rule{0.9\linewidth}{0pt}}
   \includegraphics[width=0.4\linewidth]{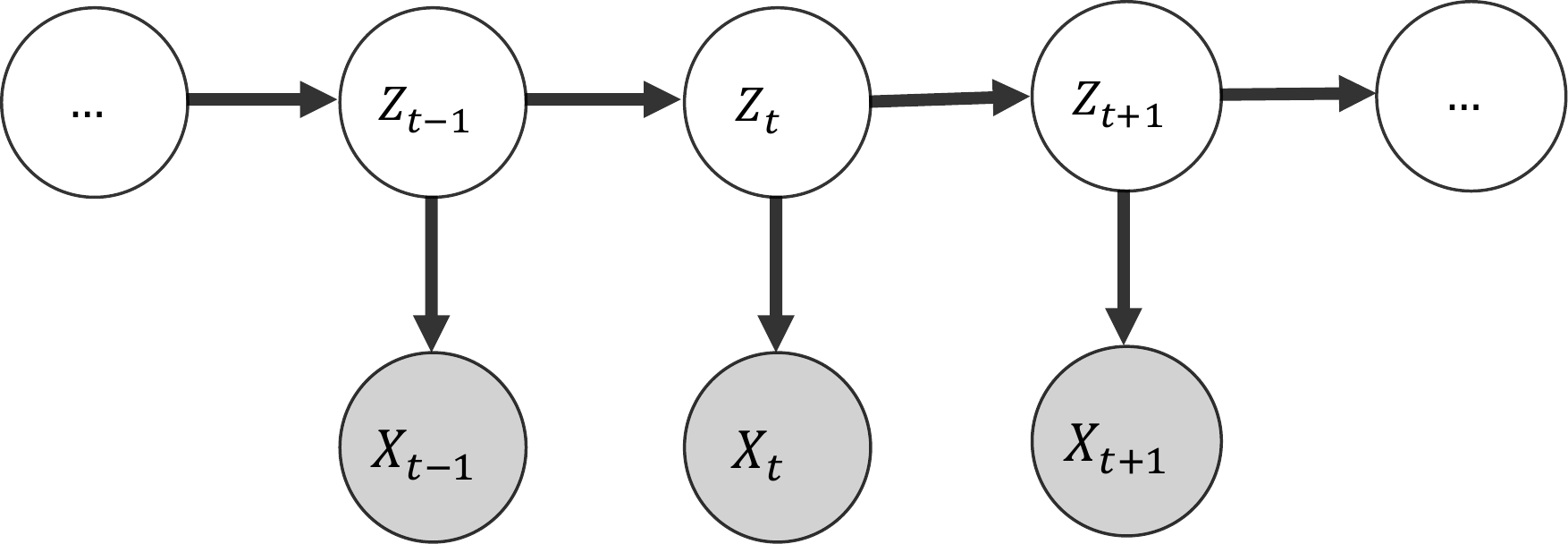}
\end{center}
   \caption{Graphical model of unimodal HMM.}
\label{fig:GM_of_hmm}
\end{figure}

A Hidden Markov Model (HMM) (\autoref{fig:GM_of_hmm}) is a doubly embedded random sequence whose underlying Markov chain is not directly observable, hence a hidden sequence. HMM model has three sets of parameters. The first set is the initial state distribution of the Markov chain, $p(\z_0)$. The second set is the transition distribution of the Markov chain, i.e. the conditional distribution $p(\z_t|\z_{t-1})$. The last set is the emission distribution $p(\x_t|\z_t)$. When $\z$ has finite discrete state space, it is a Hidden Markov Model; when $\z$ has continuous state space, it is a dynamical system. For a HMM model, we can choose arbitrary distributions, the efficient forward-backward algorithm can estimate the posterior distribution of the parameters, then use EM algorithm to improve the solution. For a review of HMM see \citep{rabiner1989tutorial}. A HMM requires a discrete state space with a known size, a dynamical system extends from finite state space to a continuous state space. A linear dynamical system simplifies a dynamical system by requiring  $\E[\z_t|\z_{t-1}]$ to be a linear function of $\z_{t-1}$. Two common forms of LDS are the Kalman filter and Kalman Smoother. To get a linear time algorithm, in a Kalman filter, we take advantage of the conjugacy of exponential family. That is,
\begin{linenomath}
\begin{align}
    p(\z_t, \{\x_{k}\}_{k=0}^{t}) = 
    p(\x_t|\z_{t})
    \int_{\z_{t-1}}p(\z_{t}|\z_{t-1})
    p(\z_{t-1}, \{\x_{k}\}_{k=0}^{t-1})
\end{align}
\end{linenomath}
which is equivalent to 
\begin{linenomath}
\begin{align}
    p(\z_t | \{\x_{k}\}_{k=0}^{t}) = 
    c_t p(\x_t|\z_{t})
    \int_{\z_{t-1}}p(\z_{t}|\z_{t-1})
    p(\z_{t-1} | \{\x_{k}\}_{k=0}^{t-1})
\end{align}
\end{linenomath}
where $c_t = p(x_t|x_1,...,x_{t-1})$ is a scaling factor.

The linear Gaussian restriction is that 
\begin{linenomath}
\begin{align}
    p(\z_t | \z_{t-1}) &\sim \mathcal{N}(\textbf{A}\z_{t-1}, \Gamma)\\
    p(\x_t | \z_t) &\sim \mathcal{N}(\textbf{C}\z_t, \Sigma).
\end{align}
\end{linenomath}
The model can be written in the conventional Kalman filter form as
\begin{linenomath}
\begin{align}
    \z_t &= \textbf{A}\z_{t-1} + \textbf{w}_t\\
    \x_t &= \textbf{C}\z_t + \textbf{v}_t\\
    \textbf{w}_t &\sim \mathcal{N}(0, \Gamma)\\
    \textbf{v}_t &\sim \mathcal{N}(0, \Sigma).
\end{align}
\end{linenomath}
Note here we have omitted the control input $\mathbf{u}$ which is assumed to be known in Kalman filter.

By recognizing the probability density function, we get the posterior distribution $p(\z_t | \{\x_{k}\}_{k=0}^{t})$.

For Kalman smoother, we have the backward form
\begin{linenomath}
\begin{align}
    p(\x_{t}, ..., \x_{T}|\z_{t-1}) 
    &= \int_{\z_t} p(\x_{t}, ..., \x_{T}, \z_t|\z_{t-1})\\
%    &= \int_{\z_t} p(\z_t | \z_{t-1}) 
%    p(\x_{t}, ..., \x_{T}|\z_{t})\\
    &= \int_{\z_t} p(\z_t | \z_{t-1}) 
    p(\x_{t}|\z_{t})
    p(\x_{t+1}, ..., \x_{T}|\z_{t}).
\end{align}
\end{linenomath}

With forward-backward filters, we can get all the statistics we need for inference and learning. For example, we can easily compute the posterior probability of $\z_t$ given the observed sequence (which is used in inference and learning)
\begin{linenomath}
\begin{align*}
    p(\z_t | \x_1, ..., \x_T) 
    &\propto p(\z_t, \x_1, ..., \x_T) \\
    &= p(\z_t, \x_1, ..., \x_t)p(\x_{t+1}, ..., \x_T |\z_t).
\end{align*}
\end{linenomath}
In speech recognition, in order to infer which word generated the sound, we need $p(\{\x_t\}_{t=1}^T | W)$ where $W$ is the word decides the probability distribution. With the forward-backwad filter, we have
\begin{linenomath}
\begin{align}
    p(\{\x_t\}_{t=1}^T | W) &
    = \int_{\z_t} p(\z_t, \x_1, ..., \x_t)p(\x_{t+1}, ..., \x_T |\z_t) \\
    &= p(\z_T, \x_1, ..., \x_T)
\end{align}
\end{linenomath}
where $\z_T$ is the final non-emitting state.

\subsection{Multimodal HMM and LDS}
\begin{figure}[t]
\begin{center}
% \fbox{\rule{0pt}{2in} \rule{0.9\linewidth}{0pt}}
   \includegraphics[width=0.4\linewidth]{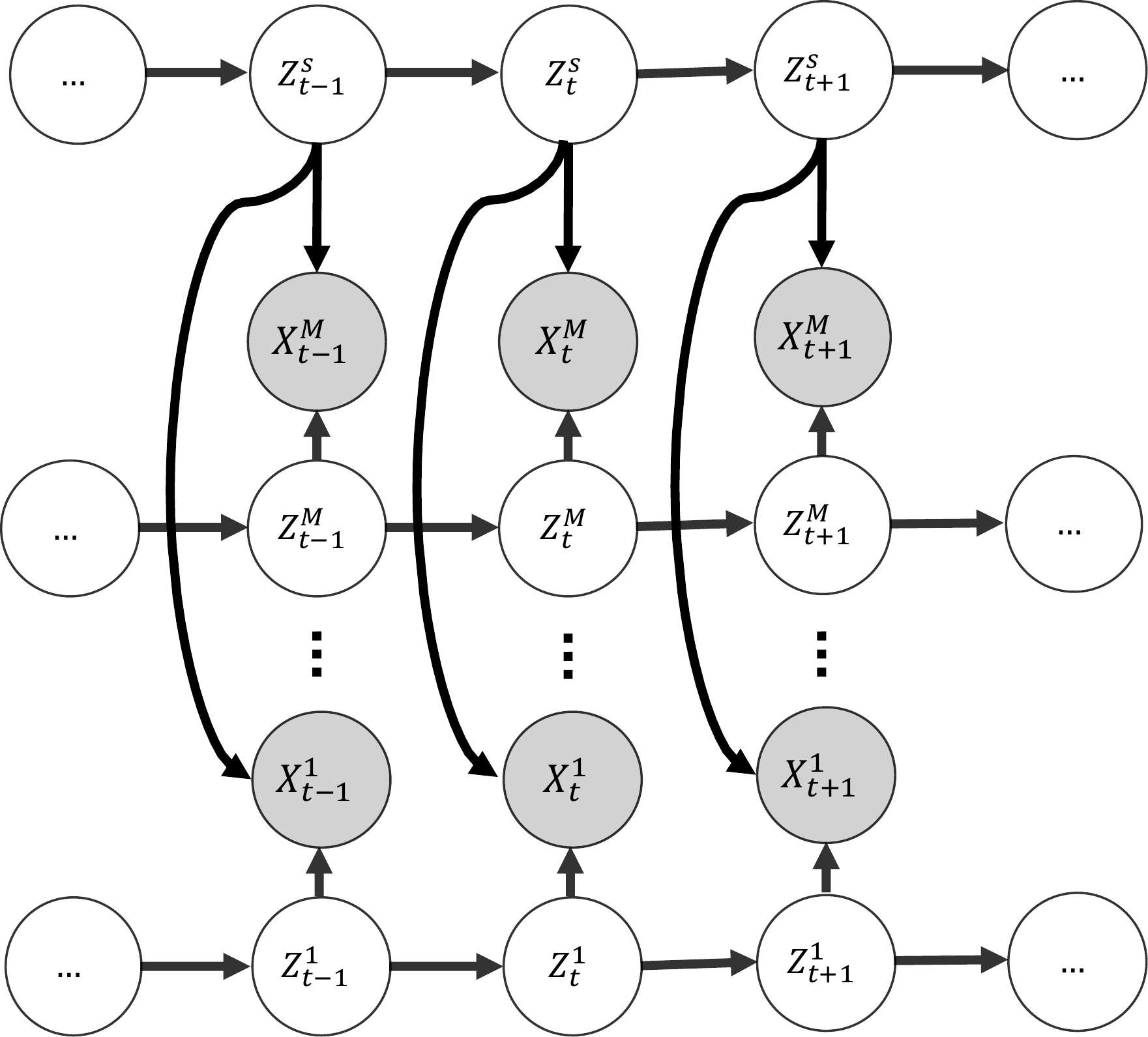}
\end{center}
   \caption{Graphical model of multimodal HMM. Each $X^m$ is a modality for $m \in \{1,...,M\}$. All modalities have the same shared latent explanatory variable $Z^s$.}
\label{fig:GM_of_sequence}
\end{figure}

First let's consider a multimodal Hidden Markov Model (HMM). Let $\{x_t^m\}_{t=1}^T$ denote the sequence of observed data of modality $m$, $\{z_t^m\}_{t=1}^T$ denote the sequence of latent explanatory factor of modality $m$, $\{z_t^s\}_{t=1}^T$ denote the sequence of latent explanatory factor shared by all modalities. The shared latent factor is a key difference between multimodal HMM and regular HMM, as it enables the observed data to be correlated between modalities. When a event (e.g. speech) is described by multiple signal sequences, the sequence of shared latent variables contains modality invariant information such as the semantic content (e.g. speech content), whereas the modality-specific latent variables contain modality-dependent information such as styles pertains only to that modality (e.g. voice timbre, face contour). We define \emph{style} latent variables to be modality-dependent,  and \emph{content} latent variables to be modality-invariant. We assume that at every time step $t$, a modality $m$ has observed variable $x_t^m$ that is generated by latent variables $\{z_t^m, z_t^s\}$. A graphical model of the sequences is shown in \autoref{fig:GM_of_sequence}. The latent variables $\{z_t^s\}$ and $\{z_t^m\}$ have Markov property such that $p(z_k^m | \{z_t^m\}_{t=1}^{k-1}) = p(z_k^m | z^m_{k-1})$ and $p(z_k^s | \{z_t^s\}_{t=1}^{k-1}) = p(z_k^s | z^s_{k-1})$. Hence at time $t$, $x_t^m$ is conditionally independent of the rest of the graph given $\{z_t^m, z_t^s\}$. Notice that the sequence $\{x_t^m\}_{t=1}^T$ is not temporally independent due to temporal dependency in $\{z_t^s\}_{t=1}^T$ and $\{z_t^m\}_{t=1}^T$. $\{x_t^m\}_{t=1}^T$ is also not spatially independent across $m$ at fixed $t$ due to the shared latent variable $z_t^s$.  We assume that modality dependent latent variables are independent, i.e. $z_k^m$ is independent of $z_k^{m'}$ if $m\neq m'$. The assumption of having orthogonal modality dependent latent factors is consistent with the objective to have disentangled latent dimensions \citep{bengio2013representation}.

Using the Markov property, we can factorize the joint likelihood in a favorable form. To simplify notation, let's denote all latent variables at time step $k$ by $\z_k = \{\{z_k^m\}_{m=1}^M, z_k^s\}$, and denote all observed variables at time step $k$ by $\x_k = \{x_k^m\}_{m=1}^M$. For a sequence of $T$ steps, the joint distribution of observed variables and latent variables is
\begin{linenomath}
\begin{align}
    \P(\x_1, ..., \x_T, \z_1, ..., \z_T) &= \prod_{t=1}^T p(\x_t | \z_{\leq t}, \x_{<t}) p(\z_t | \z_{< t}, \x_{<t})\\
%    &= \prod_{t=1}^T p(\x_t | \z_t) p(\z_t | \z_{t-1})\\
    &= \prod_{t=1}^T\left[ 
    \left(
    \prod_{m=1}^M p(x_t^m|z_t^m,z_t^s)
    \right)
    \left(
    p(z_t^s|z_{t-1}^s)\prod_{m=1}^M p(z_t^m|z_{t-1}^m)
    \right)
    \right]
\end{align}
\end{linenomath}
where the second equality follows from Markov property.

Instead of solving this multimodal HMM model, next we will present a multimodal nonlinear dynamical system, and combine it with recurrent neural network, and solve it using stochastic variational inference.

\subsection{Multimodal Variational RNN}\label{subsec-msrnn}

%\lijiang{Add discussion for why add RNN to multimodal HMM. What is the advantage of having additional RNN.}

Recurrent neural networks are able to represent long-term dependencies in sequential data by encoding inputs to update a deterministic hidden state. Recent works found that when complex sequences such as speech and music are modeled, the performance of RNN can be improved by including uncertainty in hidden states \citep{chung2015recurrent}. The reason is that when there is strong structural variation (i.e. high signal-to-noise ratio), it is mixed with random noise in inputs and outputs. While a deterministic RNN can increase its memory capacity by increasing the size of neural network, it also could bring over-fitting to training data, hence not a good solution.

The first order Markov property in HMM model is an effort to have a compromise between computation complexity and model complexity. However, the marginalization step in finding posterior distribution of $z$ is often intractable. Only a few exact inference algorithms are available (i.e. HMM and Kalman filter). 
%\lijiang{Junction tree is for which model? Markov network?}

\begin{figure}[t]
\begin{center}
   \includegraphics[width=0.5\linewidth]{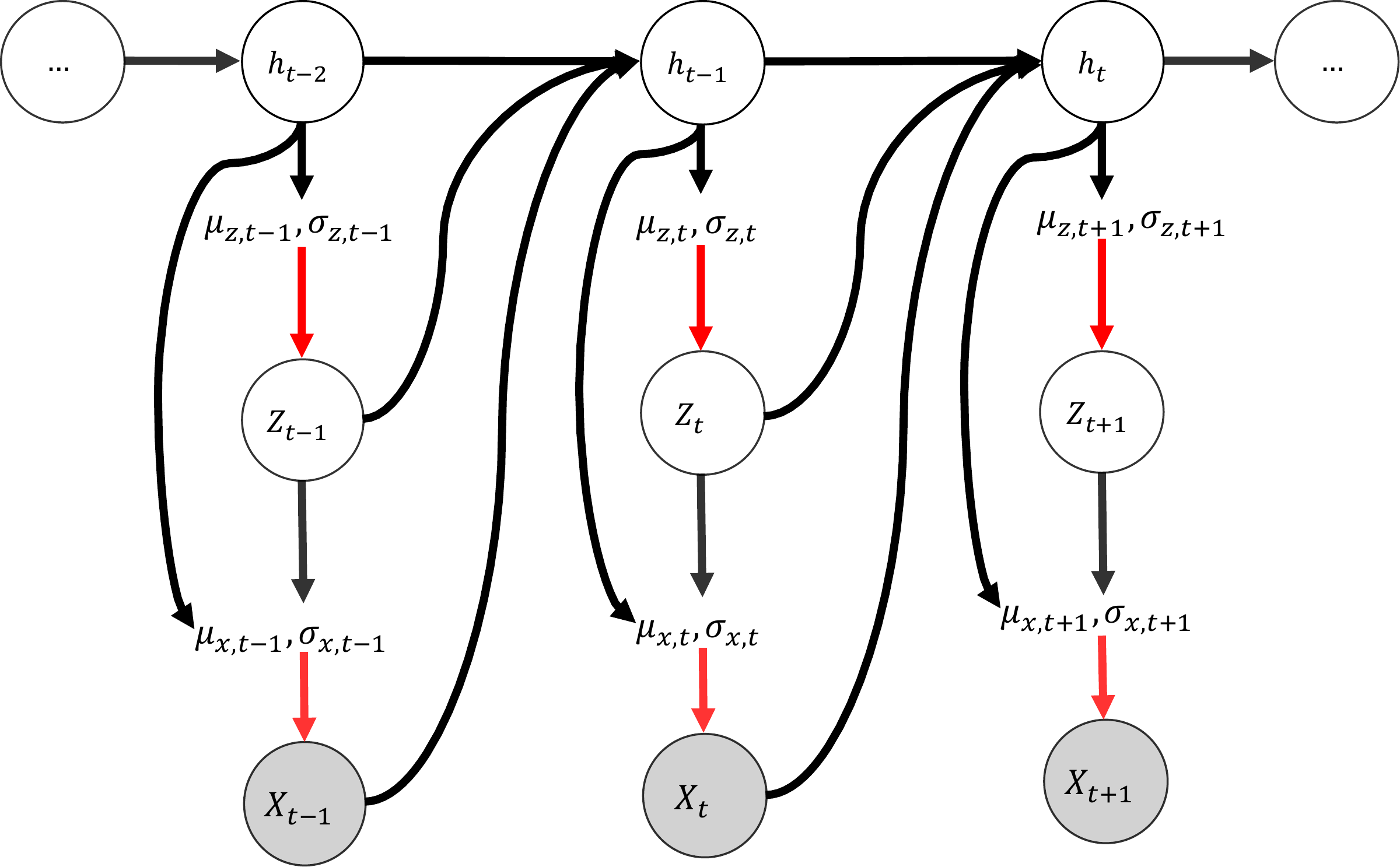}
\end{center}
   \caption{Stochastic RNN. Red line is stochastic relation, black line is deterministic relation. $h$ are hidden states of a recurrent neural network, $z$ are latent explanatory variable, $x$ are observed variables, $(\mu,\sigma)$ are parameters of Gaussian distributions.}
\label{fig:RNN-HMM}
\end{figure}

Having $p(z_t|z_{t-1}) = p(z_t|z_{t<t}, x_{<t)})$ would be ideal since it allows for maximum information passing. However, if we consider having a RNN layer on top of the latent variables as shown in \autoref{fig:RNN-HMM}, the RNN hidden state variable $h_t$ is going to pass all the previous $\z_{\leq t}$ and $\x_{\leq t}$ to timestep $t$, which breaks down the Markov property. In that case we still can factorize the joint distribution as
\begin{linenomath}
\begin{align}
    \P(\x_1, ..., \x_T, \z_1, ..., \z_T) &= \prod_{t=1}^T p(\x_t | \z_{\leq t}, \x_{<t}) p(\z_t | \z_{< t}, \x_{<t}).
\end{align}
\end{linenomath}
This long dependency is challenge to work with in dynamic Bayesian networks.  \citet{chung2015recurrent} instead used neural networks to replace the transition probability matrix and emission probability, and solve the estimation problem using stochastic variational Bayes \citep{kingma2013auto,rezende2014stochastic}.

\begin{figure}[t]
\begin{center}
   \includegraphics[width=0.6\linewidth]{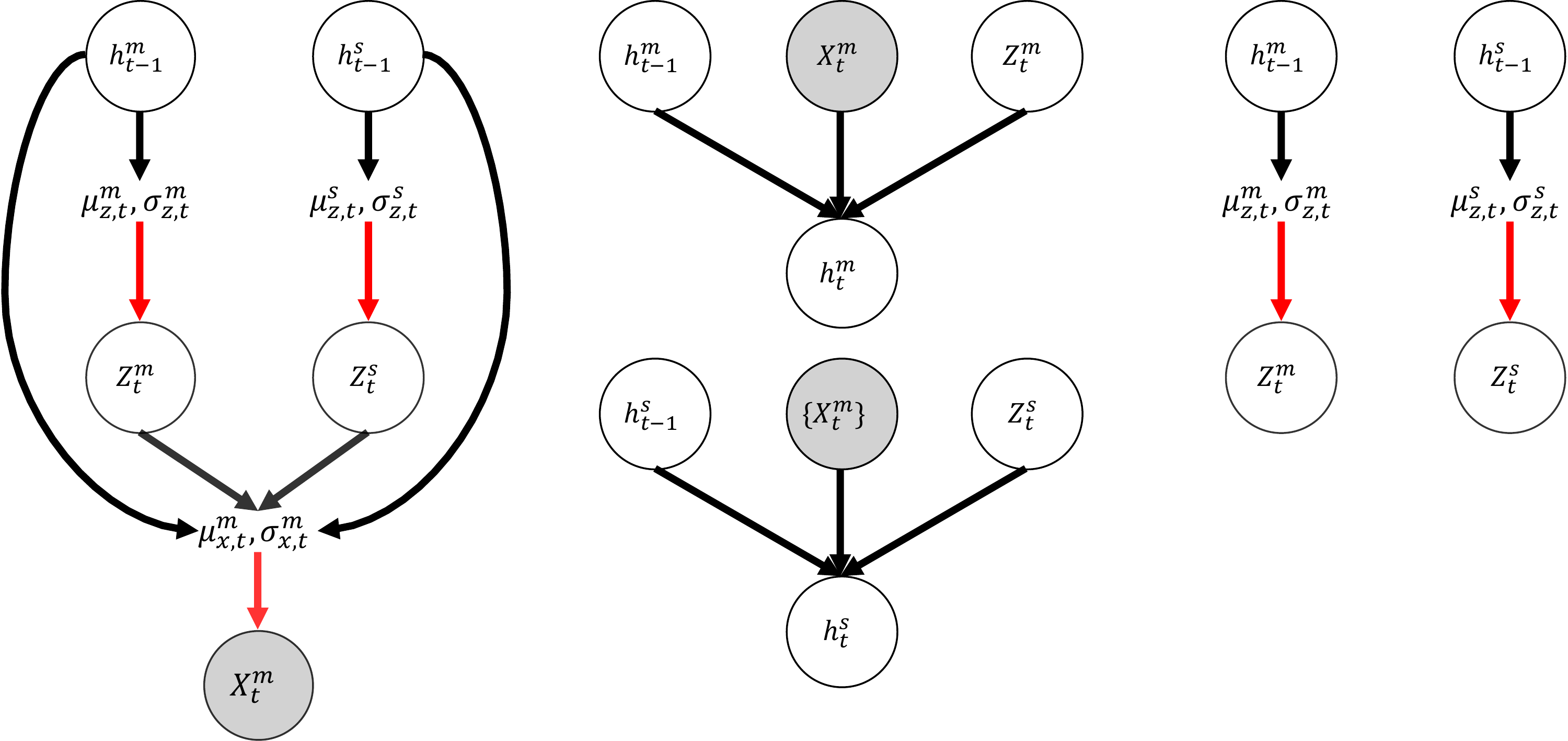}
\end{center}
   \caption{Generative model of multimodal stochastic RNN (Figure only shows one modality $m$). Red line is stochastic relation, black line is deterministic relation. $h$ are hidden states of a recurrent neural network, $Z$ are latent explanatory variable, $X$ are observed variables, $(\mu,\sigma)$ are parameters of Gaussian distributions.}
\label{fig:MVRNN-generative}
\end{figure}

In this work, we propose a multimodal variational RNN model (\autoref{fig:MVRNN-generative}). In the generative model, there are $M+1$ RNN sequences $\{\{h^m\}_{m=1}^M, h^s\}$, connecting to $M+1$ latent explanatory variable $\{\{Z^m\}_{m=1}^M, Z^s\}$. $M$ of the latent variables $\{Z^m\}_{m=1}^M$ are each associated with one modality, and $Z^s$ is shared by all modalities. As we explained earlier, each modality's output $X^m$  is a function of $Z^m$ and $Z^s$. The RNN are deterministic and the latent explanatory variables are stochastic. By doing so, we let the latent variables model the structural variation in signals, and let the RNN pass information from the past to present. The observed variables stochastically depend on latent variables, which allows for noise variations. By have two separate stochastic component, we separate structural variation from random noise variation.

\begin{figure}[t]
\begin{center}
   \includegraphics[width=0.4\linewidth]{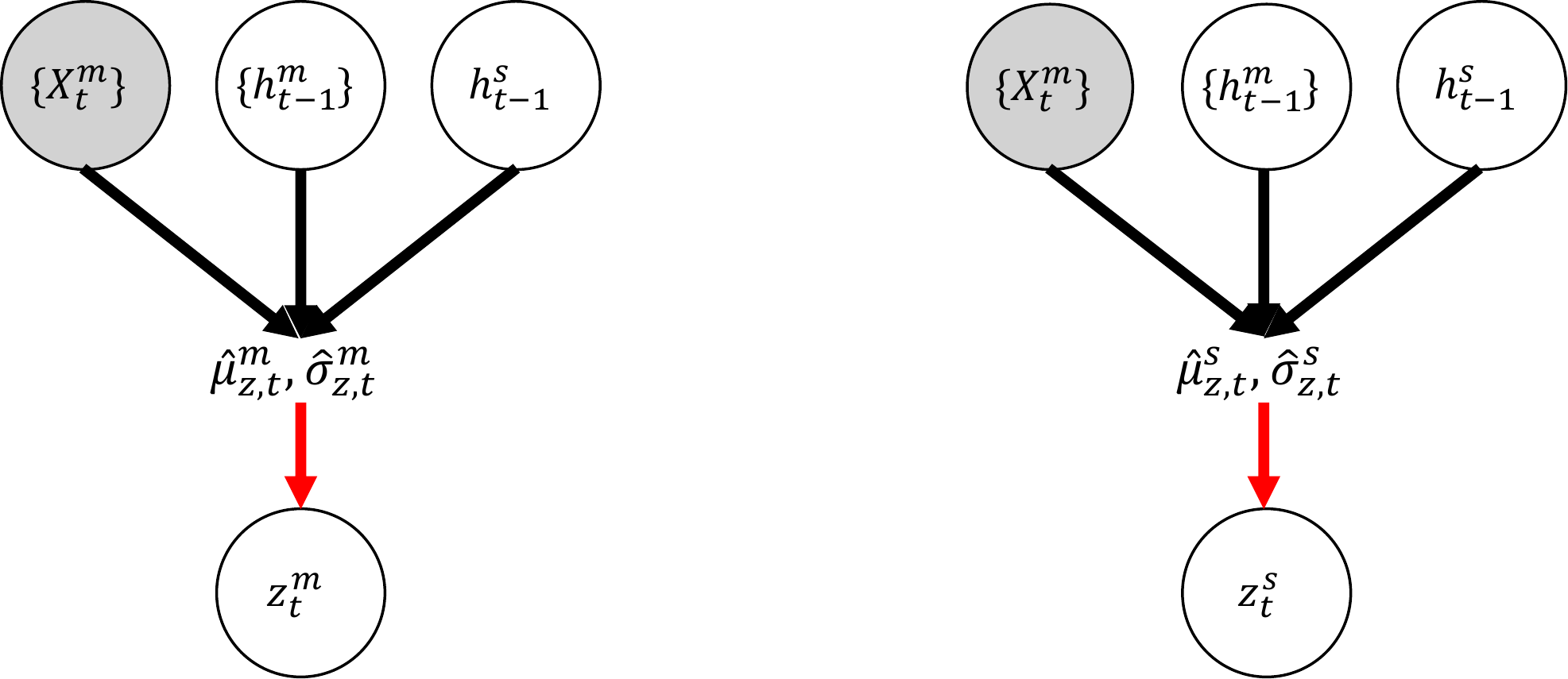}
\end{center}
   \caption{Inference Model of multimodal stochastic RNN (Figure only shows one modality $m$). Red line is stochastic relation, black line is deterministic relation. $h$ are hidden states of a recurrent neural network, $z$ are latent explanatory variable, $x$ are observed variables, $(\mu,\sigma)$ are parameters of Gaussian distributions.}
\label{fig:MVRNN-inference}
\end{figure}

For inference, we consider the following $q$ function as shown in \autoref{fig:MVRNN-inference} to approximate the posterior distribution $p(\z_1,...,\z_T | \x_1,...,\x_T)$
\begin{linenomath}
\begin{align}\label{eq:q_z}
    q(\z_1,...,\z_T) 
    &= \prod_{t=1}^T q(\z_t|\x_t, \h_{t-1}) \nonumber\\
    &= \prod_{t=1}^T q(\z_t|\z_{<t},\x_{\leq t}).
\end{align}
\end{linenomath}
The last equality holds because in the generative model (\autoref{fig:MVRNN-generative}), $\h_{t-1}$ depends on all the $\x_k$ and $\z_k$ up to time $k = t-1$. Hence our inference model (\autoref{fig:MVRNN-inference}) implies \eqref{eq:q_z}.

Thus we can derive the variational lower bound as
\begin{linenomath}
\begin{align}
    \log p(\{\x_t\}_{t=1}^T) 
%    &= \E_{q(\{\z_t\}_{t=1}^T)} \log
%    \left[ 
%    \frac{p(\{\x_t,\z_t\}_{t=1}^T}{p(\{\z_t\}_{t=1}^T|\{\x_t\}_{t=1}^T)} \frac{q(\{\z_t\}_{t=1}^T)}{q(\{\z_t\}_{t=1}^T)}
%    \right]\\
    &\geq \E_{q(\{\z_t\}_{t=1}^T)} \log
    \left[ 
    \frac{p(\{\x_t,\z_t\}_{t=1}^T}{q(\{\z_t\}_{t=1}^T)}
    \right]\\
%    &= \E_{q(\{\z_t\}_{t=1}^T)} \log
%    \frac{\prod_{t=1}^T p(\x_t|\z_{\leq t}, \x_{<t})p(\z_t|\z_{<t}, x_{<t})}
%    {\prod_{t=1}^T q(\z_t | \z_{<t}, \x_{\leq t})}\\
%    &= \E_{q(\{\z_t\}_{t=1}^T)}
%    \sum_{t=1}^T 
%    \bigg[
%    \log p(\x_t|\z_{\leq t}, \x_{<t}) 
%    + \log p(\z_t | \z_{<t}, x_{<t}) \\
%    &\quad - \log q(\z_t | \z_{<t}, \x_{\leq t})
%    \bigg]
%    \\
%    &= \E_{q(\{\z_t\}_{t=1}^T)}
%    \sum_{t=1}^T 
%    \bigg[
%    \log p(\{x_t^m\}_{m=1}^M|\z_{\leq t}, \x_{<t}) 
%    + \log p(\{\z_t^m\}_{m=1}^M | \z_{<t}, x_{<t}) \\
%    &\quad - \log q(\{\z_t^m\}_{m=1}^M | \z_{<t}, \x_{\leq t})
%    \bigg]
%    \\
%    &= \E_{q(\{\z_t\}_{t=1}^T)}
%    \sum_{t=1}^T 
%    \bigg[
%    \sum_{m=1}^M \log p(x_t^m | z_{\leq t}^m, z_{\leq t}^s, x_{< t}^m) \\
%    &\quad +
%    \sum_{m=1}^M \log p(z_t^m | z_{<t}^m, x_{<t}^m) + \log p(z_t^s|z_{<t}^s, \{x_{<t}^m\}_{m=1}^M) \\
%    &\quad -
%    \sum_{m=1}^M \log q(z_t^m | \z_{<t}, \x_{\leq t})  - \log q(z_t^s | \z_{<t},\x_{\leq t})
%    \bigg]\\
%    &= \sum_{t=1}^T \sum_{m=1}^M \E_{q(\{\z_k\}_{k=1}^t)}
%     \log p(x_t^m | z_{\leq t}^m, z_{\leq t}^s, x_{< t}^m) \\
%    &\quad +
%    \sum_{t=1}^T \sum_{m=1}^M \E_{q(\{\z_k\}_{k=1}^t)}
%    \log \frac{p(z_t^m | z_{<t}^m, x_{<t}^m)}{q(z_t^m | \z_{<t}, \x_{\leq t})}\\
%    &\quad +
%    \sum_{t=1}^T \sum_{m=1}^M \E_{q(\{\z_k\}_{k=1}^t)}
%    \log \frac{p(z_t^s|z_{<t}^s, \{x_{<t}^m\}_{m=1}^M)}{q(z_t^s | \z_{<t},\x_{\leq t})}\\
    &= \sum_{t=1}^T \sum_{m=1}^M \E_{q(\{\z_k\}_{k=1}^t)}
     \log p(x_t^m | z_{\leq t}^m, z_{\leq t}^s, x_{< t}^m) \\
    &\quad -
    \sum_{t=1}^T \sum_{m=1}^M 
    KL(q(z_t^m | \z_{<t}, \x_{\leq t})||p(z_t^m | z_{<t}^m, x_{<t}^m))\\
    &\quad -
    \sum_{t=1}^T \sum_{m=1}^M KL(q(z_t^s | \z_{<t},\x_{\leq t})||p(z_t^s|z_{<t}^s, \{x_{<t}^m\}_{m=1}^M)).
\end{align}
\end{linenomath}
Here, the first term is the reconstruction loss. The first set of KL terms is the modality specific latent variable posterior approximation error, and the second set of KL terms is the shared latent variable posterior approximation error. %A key difference between RNN-VAE and regular VAE is the the prior distribution for a latent variable $Z_t$ is not the same Gaussian$(0,I)$. Instead, the prior depends on the previous states $Z_{<t}$. Here, we define the prior distribution to be a Gaussian distribution whose parameters are given by a neural network with input $h_{t-1}$ (\autoref{fig:MVRNN-generative}).

\section{Learning Latent Variable from Similarity}\label{sec:siamese}
\subsection{Motivation}
Measuring the distance between objects has been useful in many clustering or classification tasks. People normally first find a mapping, called a embedding, of a object into a latent space, e.g. a Euclidean space. The mapping is normally hand-crafted. Here we discuss methods which learn a explicit embedding function from only knowing local similarity between sample points, not necessary distance measure, but only label of categories. Such a function can be updated as new data are collected, hence reducing the algorithm complexity. We propose a model which combines embedding and sensor fusion and discuss applications of such approach for tasks such as robust sound event detection and classification.

\subsection{Global Embedding from Local Pairwise Similarity}
Most methods for learning a latent embedding involve inverting a matrix of pairwise similarity between sample points, e.g. Locally Linear Embedding \citep{roweis2000nonlinear} and ISOMAP \citep{tenenbaum2000global}. In a high-dimensional input space, the cost of inverting such a matrix is $O(n^3)$, and often the matrix is sparse. While there are iterative methods for approximating a solution, there are other methods motivated on a objective function which can be naturally optimized using gradient descent, e.g. \emph{Stochastic Neighborhood Embedding} (SNE) \citep{hinton2002stochastic} and t-SNE \citep{van2008visualizing}. While LLE and ISOMAP lack a explicit embedding function, SNE and t-SNE models the similarity between two sample points using conditional probability, e.g. Gaussian. Suppose $x_i$ would pick all its neighbors in proportion to their probability density under a Gaussian centered at $x_i$, the similarity of sample point $x_i$ to $x_j$ is the conditional probability $p_{j|i}$:
\begin{linenomath}
\begin{align}
p_{j|i} &= \frac{\exp(-||x_i -x_j||^2 / 2\sigma_i^2)}{\sum_{k\neq i} \exp (-||x_i -x_k||^2)}.
\end{align}
\end{linenomath}
Here $\sigma_i$ is the variance of the Gaussian that is centered on $x_i$. In the embedding space (i.e. latent space), a lower dimensional counterpart $z_i$ has a similar conditional probability, which is denoted by $q_{j|i}$:
\begin{linenomath}
\begin{align}
q_{j|i} &= \frac{\exp(-||z_i -z_j||^2 / 2\sigma_i^2)}{\sum_{k\neq i} \exp (-||z_i -z_k||^2)}.
\end{align}
\end{linenomath}
If the latent points $z_i$ and $z_j$ correctly model the similarity between the high-dimensional sample points $x_i$ and $x_j$, the conditional probability $p_{j|i}$ and $q_{j|i}$ will be equal. Motivated by this observation, SNE  finds a latent representation which minimizes the difference between $p_{j|i}$ and $q_{j|i}$ for all $i,j$. SNE minimizes the sum of Kullback-Leibler divergences over all sample points using a gradient descent method:
\begin{linenomath}
\begin{align}
C = \sum_i KL(P_i || Q_i) = \sum_i \sum_j p_{j|i} \log \frac{p_{j_i}}{q_{j|i}}.
\end{align}
\end{linenomath}
One of the advantages of t-SNE is when new sample points are available, current embedding can be updated with low cost. For example, for a new point $x^*$, we can initialize its latent code $z^*$ by the mean of sample latent code, and perform gradient descent for a few iterations.

While SNE is a unsupervised learning approach, its result can be used for classification. After a embedding function $f$ is trained to convergence, it can be applied on new data for tasks involving similarity search. For example, \citet{van2008visualizing} trained a embedding function of MNIST dataset whose latent space shows good clustering results of digits. A new digit image can be send to the embedding function, and we can perform a k-nearest neighbor search to find its more probable label. Such a search is often cheaper since the embedding space is low-dimensional.

\subsection{Weakly Supervised Metric Learning without a Metric}
A drawback of SNE and t-SNE is the similarity metric is imposed with a structural form, e.g. Gaussian kernel, on both the input and latent space. \citet{hadsell2006dimensionality} proposed a \emph{siamese structure} to learn a globally coherent non-linear function that maps the data evenly in a latent space. The learning relies solely on neighborhood relationships and does not require any distance measure in the input space. 

Siamese structure is a weakly supervised model in the sense that the training data only use binary labels to learn a continuous embedding. Recognizing that a meaningful mapping maps similar input vectors to nearby points on the output space and dissimilar vectors to distant points, siamese structure constructs a contractive loss function whose minimization can produce such a embedding function. The loss function takes pairs of samples $x_i$ and $x_j$, with a binary label $y = 1$ indicating they are similar and $y = 0$ indicating they are dissimilar. In classification, samples from same classes are considered as similar, and from different classes are considered as dissimilar. The output of the function would naturally allocate each sample according to its representativeness in each class without any distance measure. Define the distance function $D_W$ as euclidean distance between the embedding output $G_W(x_1)$ and $G_W(x_2)$ where $G_W$ is the embedding function to be learned:
\begin{linenomath}
\begin{align}
D_W(x_1, x_2) = ||G_W(x_1) - G_W(x_2)||.
\end{align}
\end{linenomath}
The contrastive loss function $\mathcal{L}$ is defined as 
\begin{linenomath}
\begin{align}
L(W, (x_1, x_2, y)) &= (1-y)L_p(D_W) + y L_n(D_W),
\end{align}
\end{linenomath}
where
\begin{linenomath}
\begin{align}
\mathcal{L}(W) &= \sum_{i=1}^N L(W, (x_1, x_2, y)^{(i)})
\end{align}
\end{linenomath}
where $(x_1, x_2, y)^{(i)}$ is the $i$-th sample pair, $L_p$ is the loss function for positive pairs and $L_n$ is the loss function for negative pairs. Here $L_p$ and $L_n$ are designed to reflect similarity in the embedding space; for example, \citet{hadsell2006dimensionality} suggested using $L_p = D_W(x_1, x_2)^2$ and $L_n = \max\{0, (m - D_W(x_1, x_2))^2\}$ where $m>0$ is some hyper-parameter represents a margin for dissimilar pares.

\citet{bell2015learning}  proposed a \emph{siamese network} which utilizes deep neural network for $G_W$. The siamese network showed good result in embedding images of objects into a latent space according to their visual similarity represented by class labels.

\subsection{A Example: Audio Event Detection in Noisy Environment}
Analysis of environmental sound has the potential to be used in many applications, such as surveillance and smart homes. The Detection and Classification of Acoustic Scene and Events (DCASE) challenge is a venue for researchers to propose new methods for audio classification. Several tasks has been defined for audio classification including acoustic scene classification, sound event detection, and audio tagging. Recently, Google released an ontology and human-labeled large scale data set for audio events, i.e. Audio Set \citep{gemmeke2017audio}, which consists of 527 classes and over 2 million human-labeled 10-second long sound clips drawn from YouTube videos. Audio Set is defined for tasks such as audio tagging. The objective of audio tagging is to perform multi-label classification on fixed-length audio chunks without predicting the precise boundaries of acoustic events. Recently, deep learning methods have been successfully applied to audio tagging. %Attention mechanism \citep{xu2017attention} and gated convolutional neural network \citep{xu2018large} were used on a subset of Audio Set which contains 17 classes. \citet{kong2018audio} presented attention as a probabilistic model for the full Audio Set with 527 classes, and achieved better result than the Google's baseline (see also \citep{yu2018multi}). Other types of audio features have been used. \citet{hori2017attention} used MFCC, \citet{hori2017early} used SoundNet features \citep{aytar2016soundnet}, \citet{hori2018end} used Audio Set VGGish features \citep{hershey2017cnn}.

\begin{figure}[t!]
     \centering
     \begin{subfigure}[b]{0.45\textwidth}
         \includegraphics[width=\textwidth]{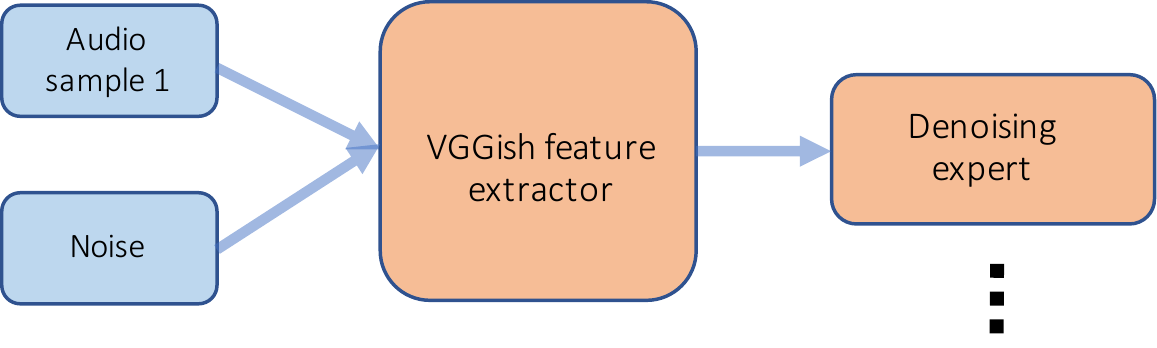}
         \caption{Step 1: Pre-train denoising experts for each noise.}
         \label{fig:dae}
     \end{subfigure}
     
     \begin{subfigure}[b]{0.6\textwidth}
         \includegraphics[width=\textwidth]{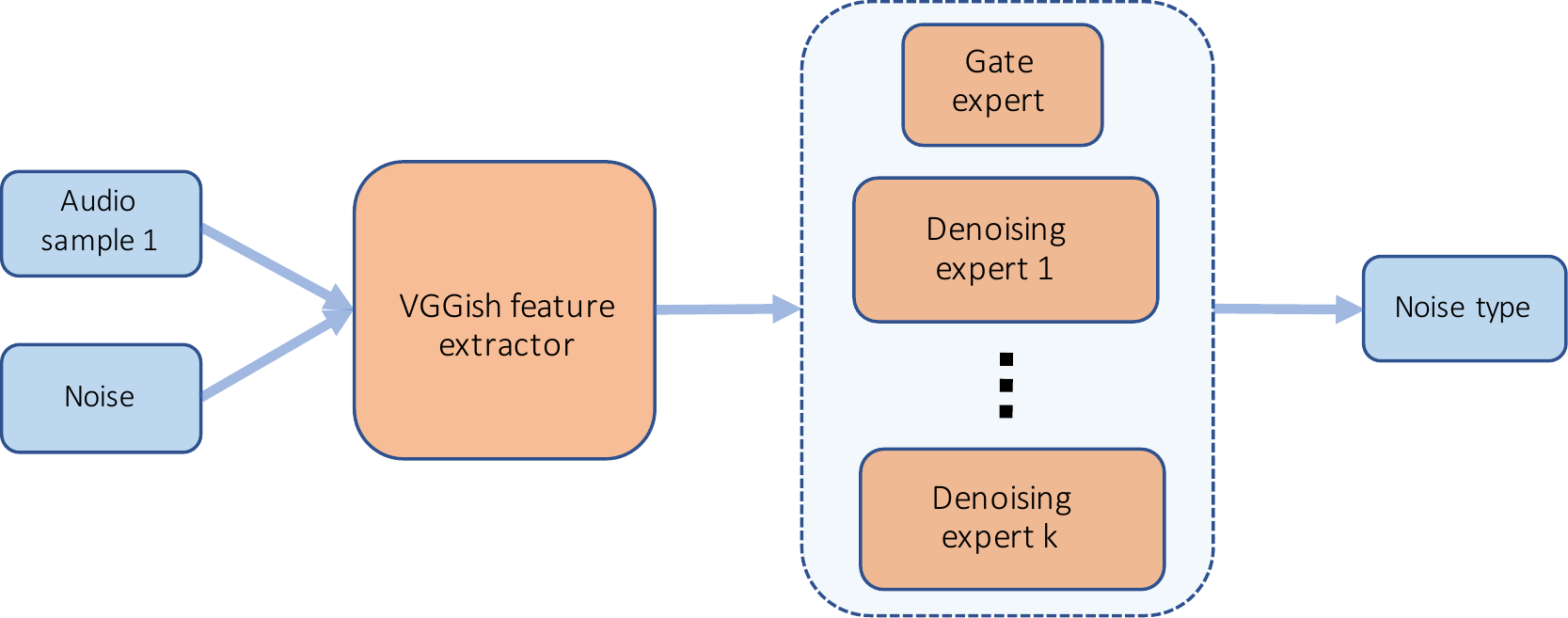}
         \caption{Step 2: Pre-train gate module for denoising expert selection.}
         \label{fig:dae-gate}
     \end{subfigure}
     
     %add desired spacing between images, e. g. ~, \quad, \qquad, \hfill etc. 
       %(or a blank line to force the subfigure onto a new line)
     \begin{subfigure}[b]{0.6 \textwidth}
         \includegraphics[width=\textwidth]{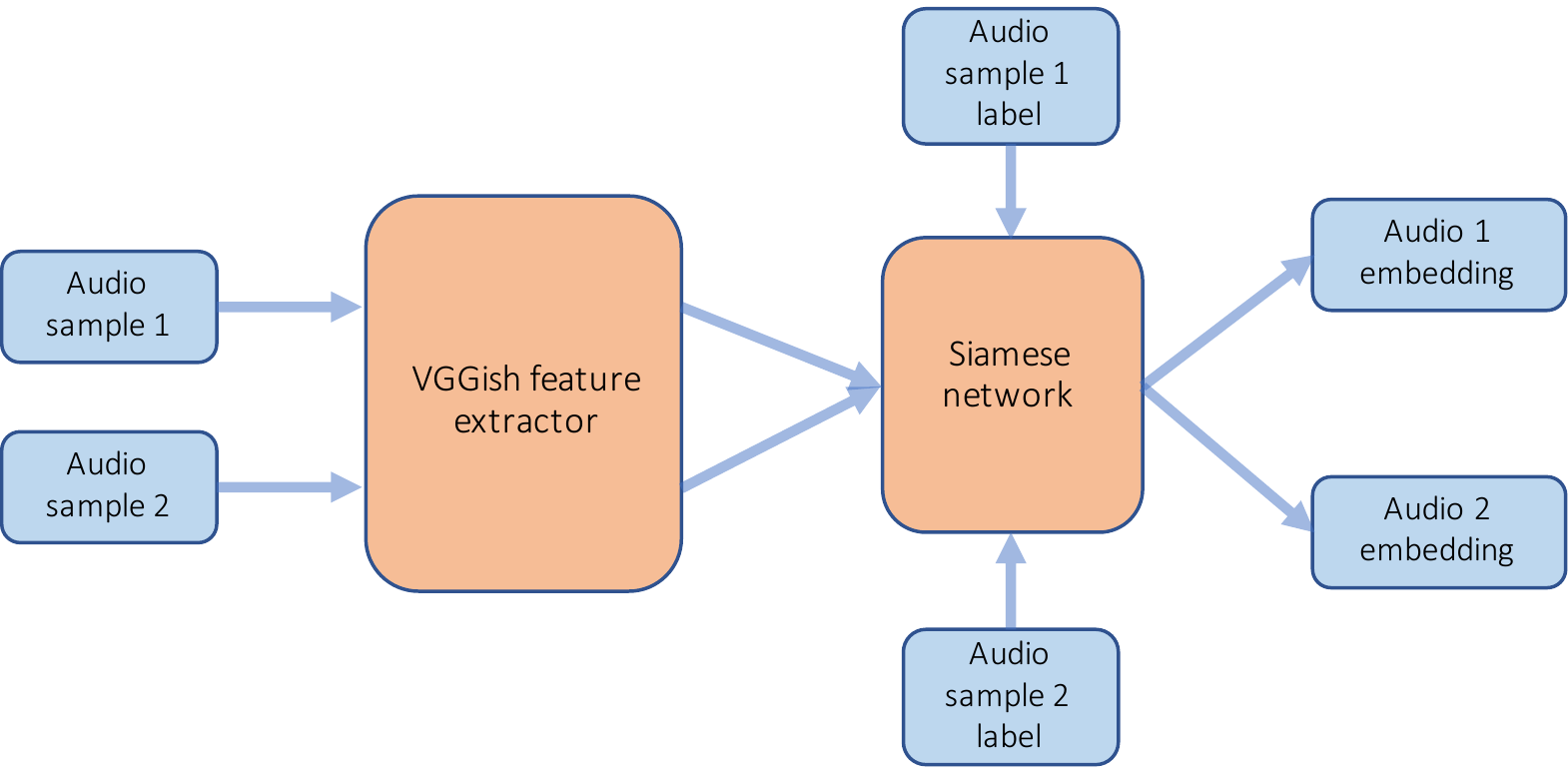}
         \caption{Step 3: Pre-train clean audio siamese network.}
         \label{fig:siamese}
     \end{subfigure}
     
     \begin{subfigure}[b]{0.9 \textwidth}
         \includegraphics[width=\textwidth]{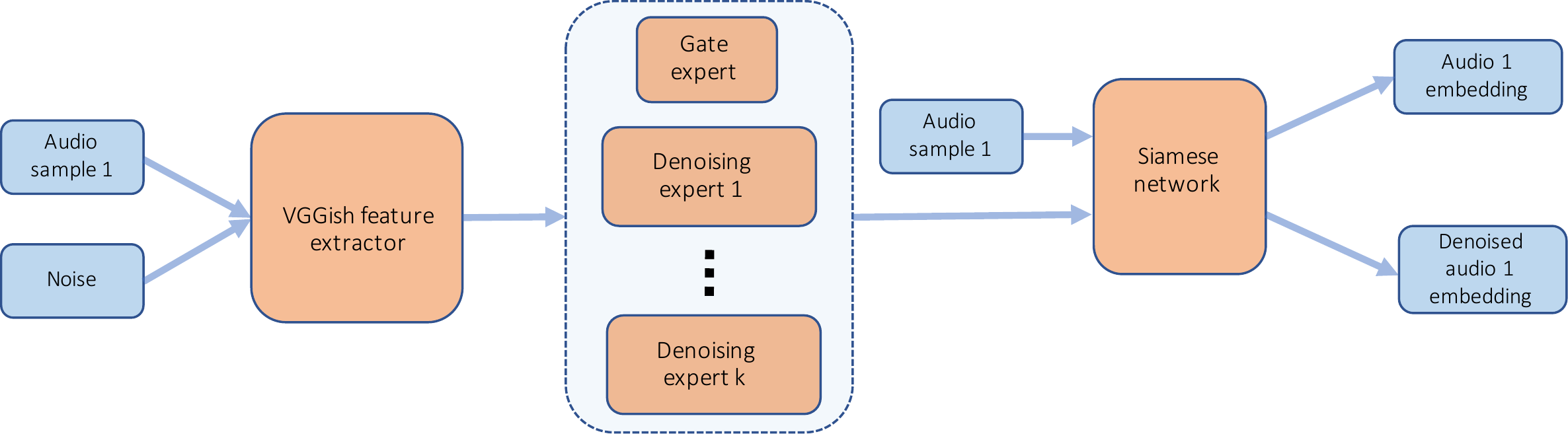}
         \caption{Step 4: Fine tune denoising expert and gate module while fix siamese network for good embedding.}
         \label{fig:siamese-dae}
     \end{subfigure}
     \caption{Robust sound event detection model.}
     \label{fig:siamese-dae-all}
 \end{figure}

Siamese network has been used for sound event detection \citep{zhang2016imisound, zhang2017iminet}. Here we explore sound event detection in a noisy environment by combining denoising autoencoder, siamese network, and conditional attention mixture model (\autoref{fig:siamese-dae-all}). We use siamese network to train a embedding encoder for a map of clean sound categories in e.g. Audio Set based on category label. Because we are not performing speech separation, we do not have to stick to STFT features as input to the network. Instead, we use pretrained features such as Audio Set VGGish features which are proved to be more effective than STFT features \citep{hershey2017cnn}. We also compare other audio features such as mel-frequency cepstral coefficient (MFCC) features and SoundNet features \citep{aytar2016soundnet}. We extract VGGish features from audios for downstream tasks as described below.

In a noisy environment, the clean sound is mixed with noise, such that the trained siamese encoder would not embed noisy sound to the exact embedding location of the clean sound. To this end, we train a denoising autoencoder (DAE) \citep{vincent2010stacked} such that the denoised sound when passed through the trained siamese encoder would map to the original clean sound embedding location as close as possible. That is, the output of the DAE is connected to the input of the siamese network to get a embedding. The loss is the difference between clean-sound embedding and denoised-sound embedding. During back-propogation, we can (1) only update the weights for the denoising network, while fixing the siamese network, or (2) only update the weights for the siamese network, while fixing the denoising network, or (3) update both. Using the siamese embedding of the denoised sound, we can classify the class of the sound event by nearest neighbor search using euclidean distance in the embedding space. This has the advantage of searching by \emph{audio similarity} \citep{zhang2017iminet}.

%For DAE, we have two options. The first option is, we train all the noises in one DAE. This would often not give as good a result as a specially trained DAE. The second option is, 
Because noise type can vary, we apply conditional attention mixture model for dynamic selection of DAE. The system is developed in four stages (\autoref{fig:siamese-dae-all}). We first pick a few kinds of noises, then pre-train one DAE for each of them (\autoref{fig:dae}). On top of pre-trained DAE's, we add the conditional attention mixture model (\autoref{eq:iid_WAL_objective}) where each DAE is a expert network, and let the gate module dynamically decide which kind of noise is most likely. This is supervised training since we know which DAE is correct during training time (\autoref{fig:dae-gate}). Meanwhile in parallel, we train a siamese network on clean audio (\autoref{fig:siamese}). After both the gate module and siamese network are trained, we connect the denoising network with the siamese network (\autoref{fig:siamese-dae}). We further fine tune the DAE's based on the distance of embedded pairs while holding siamese network as ground truth mapping. Recall the conditional attention mixture model applies a probability to each expert module, hence the choice of denoising network could be combined, and consequently the embedding into latent space could be a set of points of different weights. During test time, we can use a weighted mean of these points as the center for nearest neighbor search.

%\subsection{A Example: Multimodal Speaker Identification}
%In the first part of this report, we proposed two general multimodal learning frameworks, and demonstrated their applications in a few tasks. In last section, we proposed to combine siamese structure with conditional attention filter for audio event detection on audio inputs. It is natural to consider using siamese structure as a weakly supervised embedding algorithm for multimodal inputs, and leverage on the two co-learning methods for robustness. Here we propose a multimodal speaker identification algorithm using siamese structure and conditional attention mixture model. %In a speaker identification task, \citet{ren2016look} proposed a multimodal LSTM architecture which shares weights across time steps as well as across modalities. 

\section{Experiments}\label{sec:experiment}
We will first give a brief overview of speech activity detection and speech separation. After that we will introduce the experiment dataset and video and audio preprocessing protocol. We discuss some relevant existing deep learning techniques that can be used to improve model performance, e.g. pre-trained feature extractor. Then we explain the experiment set-up and results. Finally we discuss the experiment results and propose follow-up works to be completed in our follow up works. Due to time constraints, we are not able to complete all the works we hoped at this time. We plan to complete them as part of the dissertation thesis.

\subsection{Speech Activity Detection and Speech Separation}
Acoustic event detection (AED) has draw many attentions due to its wide application in intelligent systems such as robots and smart homes \citep{gemmeke2017audio, hershey2017cnn}. With new tools such as deep learning, we have seen significant improvement in the results in recent DCASE Acoustic Scene Classification (ADC) task. As one kind of AED tasks, Speech activity detection (SAD) is a classification problem of a given sequence of audio frames into speech active and non-active states. Most SAD models rely on audio signal, for example \citet{li2017endpoint} used grid LSTM to detect speech endpoints. Although speech is audio signal, video signal has shown value to detection and understanding of speech.  Sadly, most SAD systems developed so far either entirely rely on audio or video alone, only a limited number of systems utilize both audio and video signal, e.g. \citep{ariav2018deep}. In addition, it is not a straightforward task to effectively combine audio and video signals due to the natural differences between audio and video signals. 

\begin{figure}[t!]
\begin{center}
% \fbox{\rule{0pt}{2in} \rule{0.9\linewidth}{0pt}}
   \includegraphics[width=0.9\linewidth]{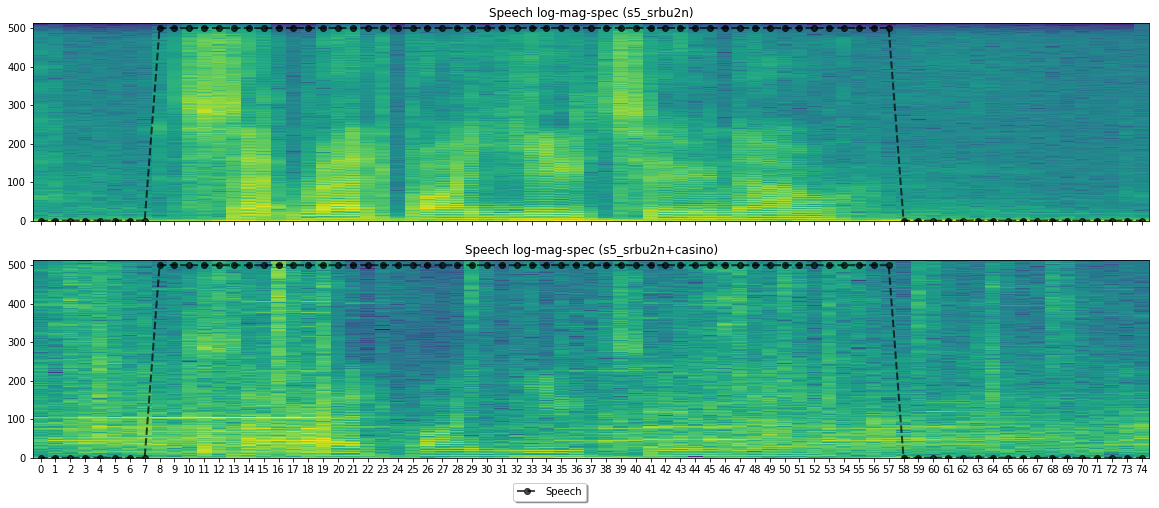}
\end{center}
   \caption{Log-mag-STFT of speech. Top figure shows speech, bottom figure shows speech plus casino noise with SNR = -5DB. Dotted black line shows speech activity.}
% \label{fig:long}
\label{fig:log_mag_STFT_speech_casino5}
\end{figure}

A unimodal system that relies solely on the audio signals can fail to do this job due to the common artifacts found in the real-world speech signals such as additive noise and reverberation. To give a concrete example, one speech audio is displayed as log transformed magnitude Short Time Fourier Transform (log-mag-STFT) in \autoref{fig:log_mag_STFT_speech_casino5}. Notice although the speech only appears between 8th frame and 57th frame, it is difficult to determine where are true speech activities due to the noise we manually injected. In this scenario, video provides a opportunity to improve the accuracy of speech activity detection. 

\begin{figure}[ht!]
\begin{center}
% \fbox{\rule{0pt}{2in} \rule{0.9\linewidth}{0pt}}
   \includegraphics[width=0.5\linewidth]{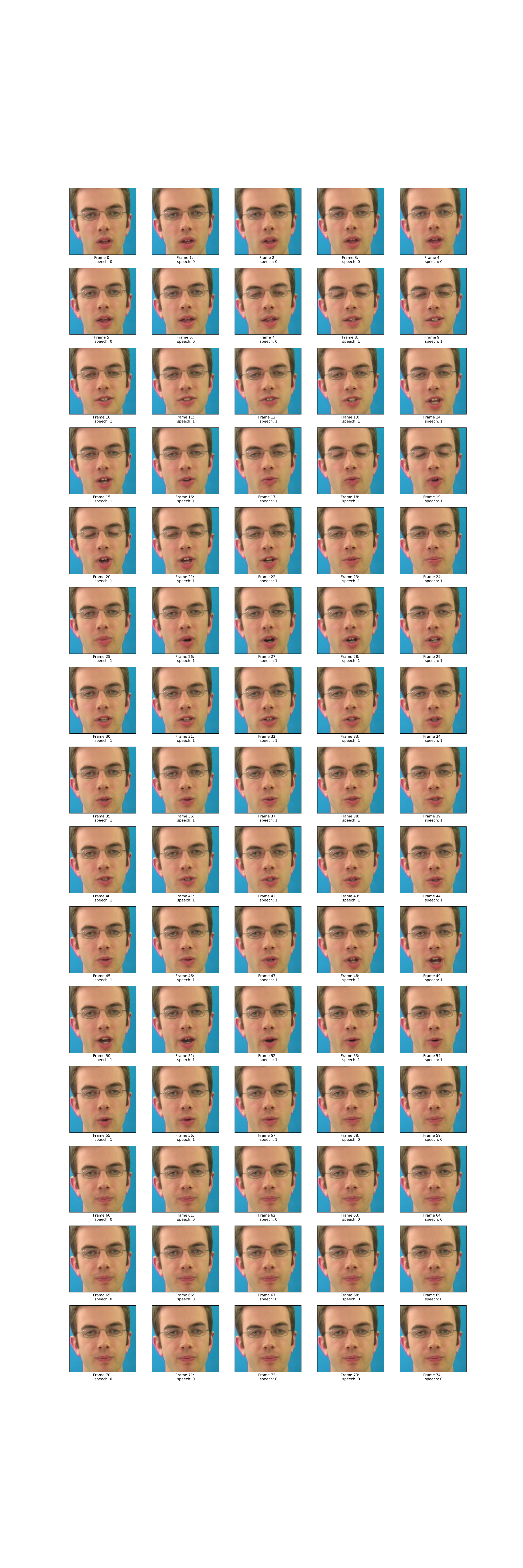}
\end{center}
   \caption{Images of speech (3 seconds, 25fps). Speech activity is from frame 8 to frame 57.}
% \label{fig:long}
\label{fig:faces_s5_srbu2n}
\end{figure}

While video signal is invariant to acoustic environments, the relation between speech and face movement is dynamic and usually asynchronous. For example, when uttering some words, part of the lip movements (i.e. visemes) are loosely coordinated with the sound (i.e. phoneme). In this sense, two input modalities are not independent. Visually speaking, speech is closely related to mouth movements. Hence the visual speech activity detection problem can be easily contaminated by non-speech mouth movement such as breathing, eating, etc. For example, in \autoref{fig:faces_s5_srbu2n}, speech starts at 8th frame, but there is noticeable mouth movement from frame 0 to 7. Phonetically speaking, vowel is a speech sound tends to require relatively open mouth, while a consonant is a sound made with mouth relatively closed. Hence speech does not necessarily activates expressive mouth movements, which puts challenge on video-only speech activity detection. More importantly, visual signals do not always provide a reliable cue due to variations such as head rotations, illuminations, different view points, etc.

We want to comment that video based Automatic Speech Recognition (ASR) has been addressed in a few previous researches; for example \cite{assael2016lipnet}. These models directly classify video segments into words or phonemes, and hence can be used for SAD. We approach the SAD problem from a different perspective. We consider ASR as a multi-stage system, and SAD is a key step in front end processing. Instead of using video to predict text, we use visual cue to assist audio cue to classify noise audio from noisy speech audio. When noisy speech audio can be located (e.g. black line in \autoref{fig:log_mag_STFT_speech_casino5}), we can apply source separation methods to extract clean speech from noisy speech. The denoised speech is subsequently fed into a audio based ASR system. To this end, video based ASR can work with audio based ASR in our model jointly in a multimodal ASR task.

A SAD system can be applied to convert a unsupervised speech separation systems into a supervised one. Consider a noisy acoustic environment, if the type of noise is unknown, we have a unsupervised speech separation problem. If we know the type of noise, we have a supervised speech separation problem, which is significantly less challenging than the unsupervised case. With a SAD system, we can detect the noisy period immediately before speech. If we assume that the same type of noise will continue during speech, then we can use dictionary based speech separation algorithms such as \citep{smaragdis2007supervised} with a known dictionary, hence a supervised speech separation system. In a multi-person speech separation scenario, SAD system can assist a face recognition algorithm to determine which person is speaking, and use beam-forming to enhance that speaker's speech.

We emphasize again it is easy to modify our model for speech activity detection to speech separation. For speech separation, a common technique is to use a ideal ratio mask (or a ideal binary mask), which is a element-wise ratio (or binarized ratio) between clean and noisy spectrogram. This mask is then multiplied with the input spectrogram to get a denoised spectrogram. The audio input could be either complex spectrogram or magnitude spectrogram. In the case of complex spectrogram, two masks will be generated for real and imaginary component respectively. When using magnitude spectrogram, one mask will be generated, and the phase of noisy input will be used to as denoised phase. The speech activity detection model discussed above can be seamlessly transformed into a speech separation model.

\subsection{Data}

 \begin{figure}[t!]
     \centering
     \begin{subfigure}[b]{0.46\textwidth}
         \includegraphics[width=\textwidth]{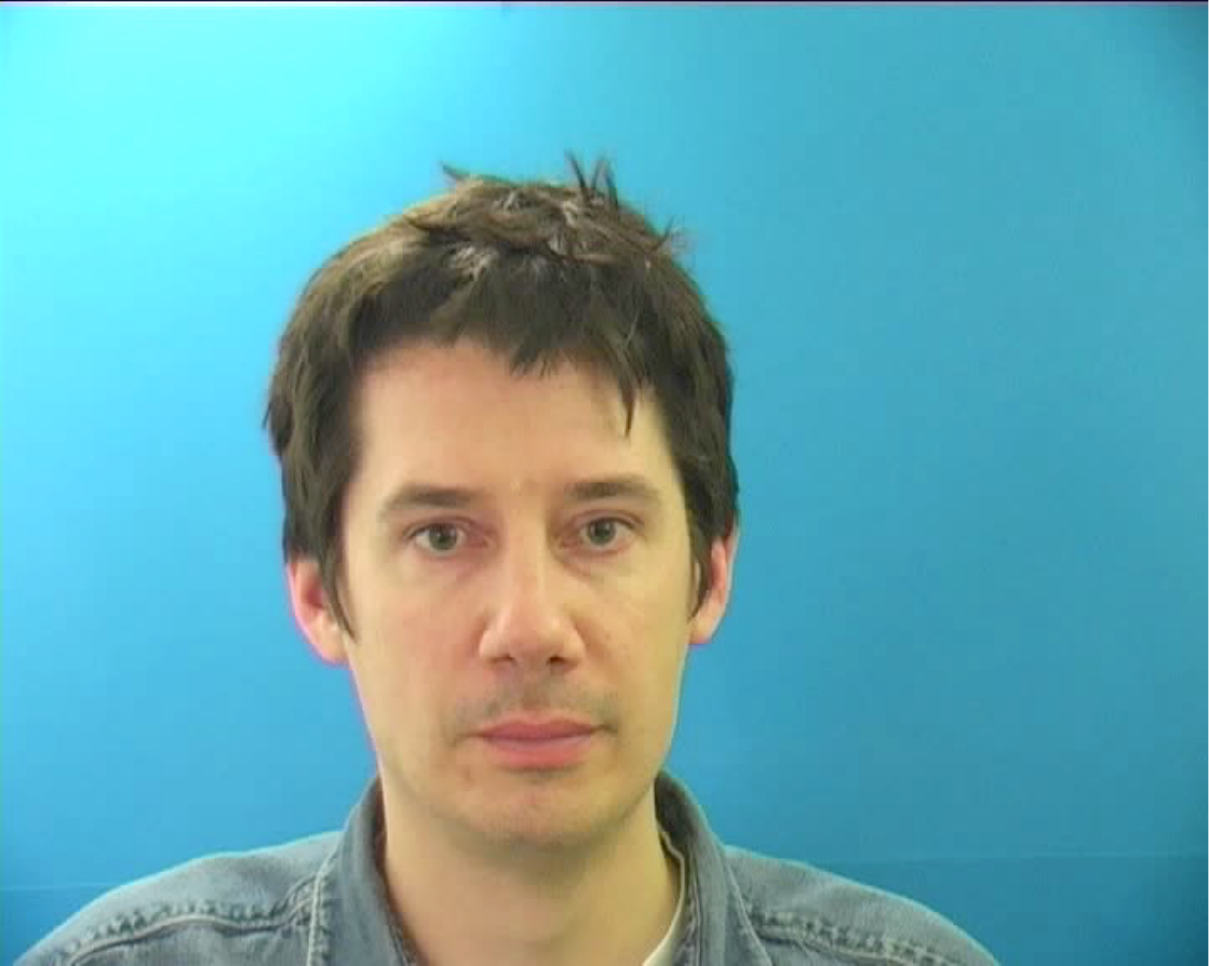}
         \caption{One frame of video image.}
         \label{fig:grid-p1}
     \end{subfigure}
     \qquad
     %~ %add desired spacing between images, e. g. ~, \quad, \qquad, \hfill etc. 
       %(or a blank line to force the subfigure onto a new line)
     \begin{subfigure}[b]{0.37\textwidth}
         \includegraphics[width=\textwidth]{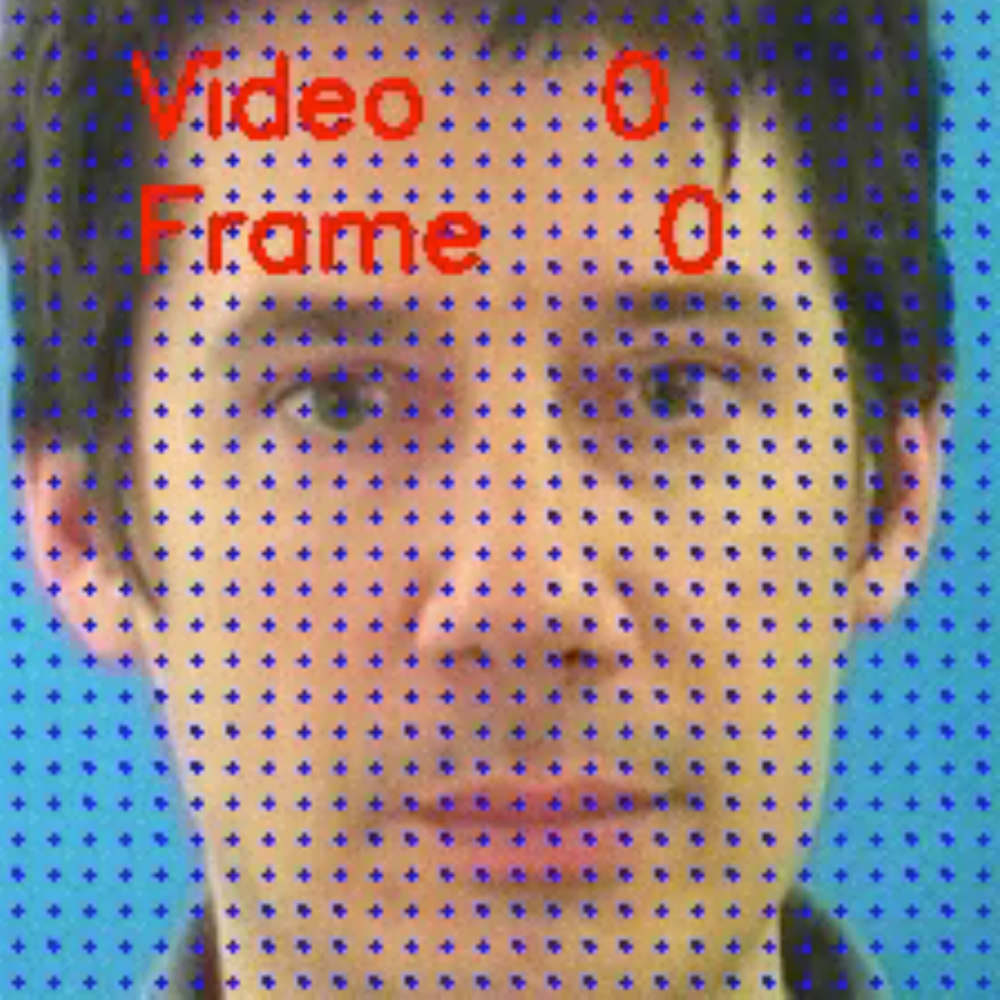}
         \caption{Face image (with optical flow).}
         \label{fig:grid-p1-face}
     \end{subfigure}
     \caption{Video image of GRID.}
     \label{fig:grid-p1}
 \end{figure}

In this experiment we use the GRID corpus data \citep{cooke2006audio} which contains 34 speakers. Each speaker has 1000 short speeches, each about 3 seconds long, recorded in a controlled environment with limited ambient noise. The speech is fully annotated, with phonemes mapped to time. \autoref{fig:grid-p1} shows a sample image of one video. 

Due to limitation in computing resources, in this pilot study we randomly choose a subset of subjects to develop our model. For the experiment results presented in the following sections, we use 4 subjects, 2 males and 2 females, to train and test our model. For each subject, we randomly partition the video clips by $7:2:1$ into training, validation, and in-sample testing sets. We have a 5th subject served as out-of-sample testing set. We are planning on extend to all subjects in  following works.

\subsection{Video Pre-processing and Pre-training}
Each video clip is 3 seconds long with 25 fps. We pre-process the video data using \verb|opencv-python|. However, due to \verb|ffmpeg| issue, some unpacked videos have less than 75 frames. To align the video with audio, we pad the first and last frames to the front and end of video to increase the number of frames to 75.

In visual activity detection, still image and motion between frames have been shown to complement each other \citep{simonyan2014two}. Therefore, we extract two kinds of information from video: image and motion. We first extract a face from each video frame. Because the video is relatively clean with speaker's front face at the center the video, we found a Viola-Jones type face detector works well. We used \verb|opencv-python| Harr cascade front face feature to detect faces. We are able to detect high quality face bounding box because the camera is directly pointing at face of the subject, and both camera and subject are stationary. However, there is still movements of subjects between frames, such that each bounding box shifts a little. Because the subject in each video is always stationary and the movement is not dramatic, we choose to use the bounding box in first frame for the 75 frames. Each face image is resized to $224 \times 224$ pixels in order to feed into a InceptionV3 feature extraction.

To create image features, pre-trained feature extractors can be used such as AlexNet \citep{krizhevsky2012imagenet}, VGG \citep{simonyan2014very}, InceptionV3 \citep{szegedy2016rethinking} and ResNet-50 \citep{he2016deep}. In this experiment, we use InceptionV3 pre-trained on ImageNet without fine-tuning. The extracted InceptionV3 feature has dimension of 2048 for each $224\times224\times3$ image input. Alternatively, with enough data and computing resource, we could also train the feature extractor from scratch on GRID images or a general human face dataset, and fine-tune with the rest of the model. We will pursue these works in future.

Convolutional network trained on multi-frame dense optical flow has been shown to achieve good performance on capturing human motion \citep{simonyan2014two}. We compute dense optical flow using Farnback algorithm provided in \verb|opencv-python| from the extracted face images. Optical flow is $224 \times 224 \times 2$ for each frame.

\begin{figure}[!ht]
\begin{center}
% \fbox{\rule{0pt}{2in} \rule{0.9\linewidth}{0pt}}
   \includegraphics[width=1\linewidth]{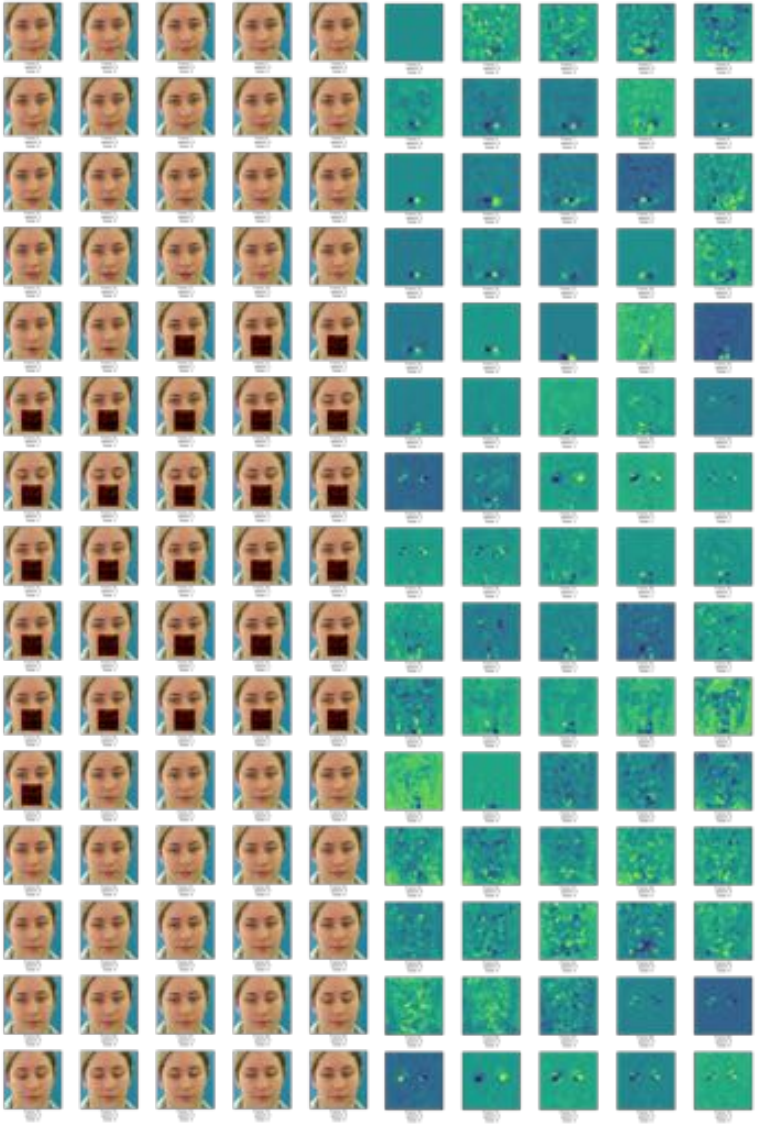}
\end{center}
   \caption{Corrupted images and optical of speech (subject 11, speech srbt7s, 3 seconds, 25fps). Speech activity between 9th and 52nd frames, video noise between 22nd and 49th frames.}
% \label{fig:long}
\label{fig:corrupted_face}
\end{figure}

To create a corrupted video signal, we randomly apply a square masking patch, and randomly adjust the brightness of that patch area. The location of the patch is random for each video, and the size of the patch is randomly chosen between 50 and 150. For each video, we randomly pick a segment of 20 to 30 frames, and apply the patch to all frames in that segment. This mimics the effect of having a shadow or spot-light casting on the subject's face during a segment of a video clip. \autoref{fig:corrupted_face} shows a example of corrupted image and corrupted optical flow of the same video.

\subsection{Audio Pre-processing and Pre-training}
\begin{figure}[!ht]
\begin{center}
% \fbox{\rule{0pt}{2in} \rule{0.9\linewidth}{0pt}}
   \includegraphics[width=1\linewidth]{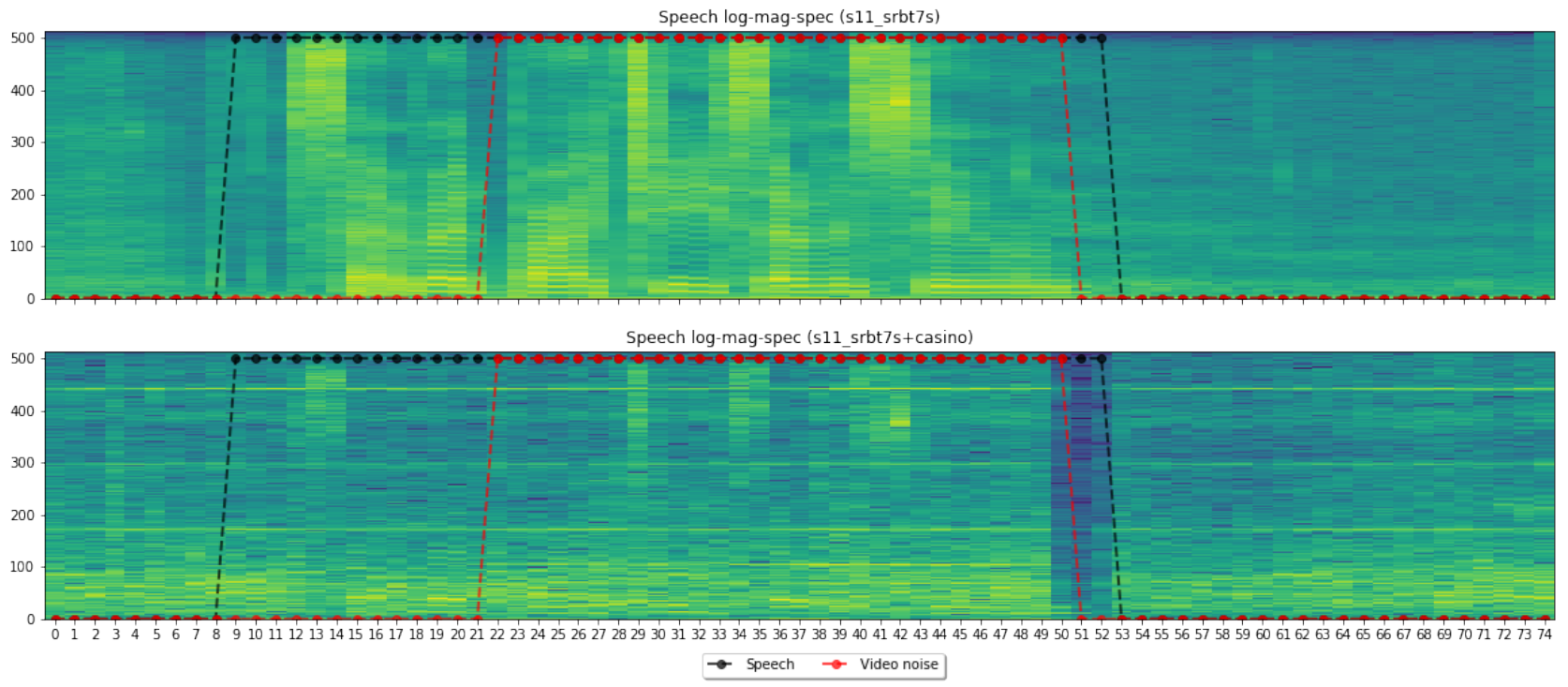}
\end{center}
   \caption{A example of corrupted speech. Top clean speech, bottom corrupted speech (subject 11, speech srbt7s, -5db SNR with Casino noise). Black line marks speech activity, red line marks visual noise (audio noise is presented in all frames).}
% \label{fig:long}
\label{fig:p11-srbt7s}
\end{figure}

For audio data, we first pre-process the input audio to synchronize with images extracted from video by matching number of audio frames with the number of image frames. We choose the STFT window size and re-sample audio such that the there are 25 frames per second for audio. We first down-sampled the speech audio from 50kHz to 16kHz, then perform Short-Time-Fourier-Transform with a window size of 1024 and a stride of 760 which transforms each audio into $75 \times 513$ representation. Notice that 25 frames per second may not be sufficient to fully capture the dynamics of speech. Our method can be easily extended to extract 50 or 75 audio frames per second.

Similar to image feature extractor, we consider audio features such as mel-frequency cepstral coefficient (MFCC) features, SoundNet features \citep{aytar2016soundnet} and Audio Set VGGish \citep{hershey2017cnn}. Because MFCC and VGGish features are not invertible, in speech separation, we use STFT features as input to the network. In acoustic event detection, MFCC, SoundNet and VGGish are proved to be more effective than STFT features \citep{hershey2017cnn}. We will pursue these works in future.

To create a corrupted speech signal, we injected two kinds of noise to the audio. The first kind is a casino noise proposed by \citep{duan2012speech}. The casino noise is challenging due to being a non-stationary, wide-band noise which span from low to high frequency domain. \autoref{fig:p11-srbt7s} shows a example.

We construct a second kind of noise using a mixture of three person from TIMIT training data (TIMIT-3P). TIMIT training set contains 136 female and 326 male speakers, while the testing set contains 56 female and 112 male speakers, which are from eight dialect regions in the US. Each TIMIT speaker has 10 short utterances. For each GRID speech, we randomly choose 3 speeches from TIMIT, then choose a random segment of 3 seconds from each chosen speech. We mix the 3 speeches with equal magnitude. Finally, we mix the 3-person mixture with GRID speech with SNR equal 0 DB.

%\subsection{Adversarial Training}
%Adversarial autoencoder

\subsection{Training}
During each training epoc, a minibatch of size 16 is used. Same minibatch size is used for evaluating validation and testing set. For RNN we used a 5 frames for truncated backpropagation through time. For DNN model, we feed 5 audio frames per sample to provide a context, which is a common practice in audio based speech recognition system. All models are trained on a server with 4 Nvidia Titan X GPUs with 12GB RAM per GPU. The multimodal RNN model takes about 1-4 hours \footnote{It seems the training speed is fluctuating on the server for some unknown reason.} per training epoc, and we trained it for three days of 30 epocs. The recurrent attention mixture model takes about 4 hours per epoc to train . 

\subsection{Experiment Results}
Our first set of experiments is on the recurrent attention filter. We implemented four versions of the model. The first one is the conditional attention mixture model (\autoref{eq:iid_WAL_objective}). The second one is the recurrent attention mixture model (\autoref{eq:WAL_objective}). For each of these two models, we construct two versions, i.e. with and without co-learning (distance based regularizer). For comparison, we also implemented three uni-modal DNN models (audio, optical flow, and image). The results are shown in \autoref{table:result}.

First we notice that multimodal input model does outperform unimodal input model. While the no-audio models (i.e. image and optical flow) have lower accuracies than the audio-only model, most multimodal input models outperform audio-only model, which suggests image and motion provides addition information not seen in audio for speech activity detection. This validates our hypothesis that multimodal input is useful to speech related tasks. 

Second, we notice that the RNN model easily overfits the training data. While the recurrent attention model has lower training error than the conditional attention model, during test time the result is reversed. This suggests we may need (1) to regularize RNN such using e.g. dropout or (2) stop RNN training earlier. 

Third, we notice that a simple distance based regularizer for co-learning does not improve prediction accuracy. On the contrary, the un-regularized models have higher prediction accuracy, very likely due to the lack of restriction of the model. Recall our motivation for the model-based regularization is that co-learning helps to separate modality invariant features from modality-dependent features, however, such features may not be useful for speech activity detection. This suggests we may need to carefully design the co-learning structure in order to extract features that are useful to a task.

\begin{table}[t]
\centering
\caption{Experiment Results}
\label{table:result}
\resizebox{0.6\columnwidth}{!}{
 \begin{tabular}{ |l||l|l|}
 \hline
  & \textbf{Train(*)} 
  & \textbf{Test (+)}\\
 \hline
 \hline
 \multicolumn{3}{|c|}{\textbf{Uni-modal input}} \\
 \hline
 Image   & 0.12 & 80.8 \\
 Optical flow & 0.03
 & 90.5 \\
 Audio (with casino noise)   & 0.06 & 92.5 \\
 Audio (with 3 persons noise)   & 0.06 & 91.1 \\
 \hline
 \hline
 \multicolumn{3}{|c|}{\textbf{Multimodal input without co-learning}} \\
 \hline
Conditional Attention Mixture (with casino noise)   & 0.03 & 96.38 \\
Conditional Attention Mixture (with TIMIT-3P noise)   & 0.04 & 96.25 \\
Recurrent Attention Mixture (with casino noise)   & 0.03 & 93.75 \\
Recurrent Attention Mixture (with TIMIT-3P noise)   & 0.1 & 85.7 \\
 \hline
 \hline
 \multicolumn{3}{|c|}{\textbf{Multimodal input with distance based co-learning}} \\
 \hline
Conditional Attention Mixture (with casino noise)   & 0.03 & 91.74 \\
Conditional Attention Mixture (with TIMIT-3P noise)   & 0.1 & 84.56 \\
Recurrent Attention Mixture (with casino noise)   & 0.03 & 89.37 \\
Recurrent Attention Mixture (with TIMIT-3P noise)   & 0.1 & 76.23 \\
 \hline
 \hline
 \multicolumn{3}{|l|}{\text{*: MSE, +: \% correct.}} \\
%  \hline
%  \multicolumn{3}{|l|}{\text{-:The multimodal RNN occupies all my GPUs (4 Titan X), and takes very long time to train, I'm waiting for its and other results.}} \\
 \hline
\end{tabular}}
\end{table}

% \texttt{I found that I am missing speaker S2's $timit_3p_snr0$ speech data. I'm worried it may have cause the }

 \begin{figure}[t!]
     \centering
     \begin{subfigure}[b]{\textwidth}
         \includegraphics[width=\textwidth]{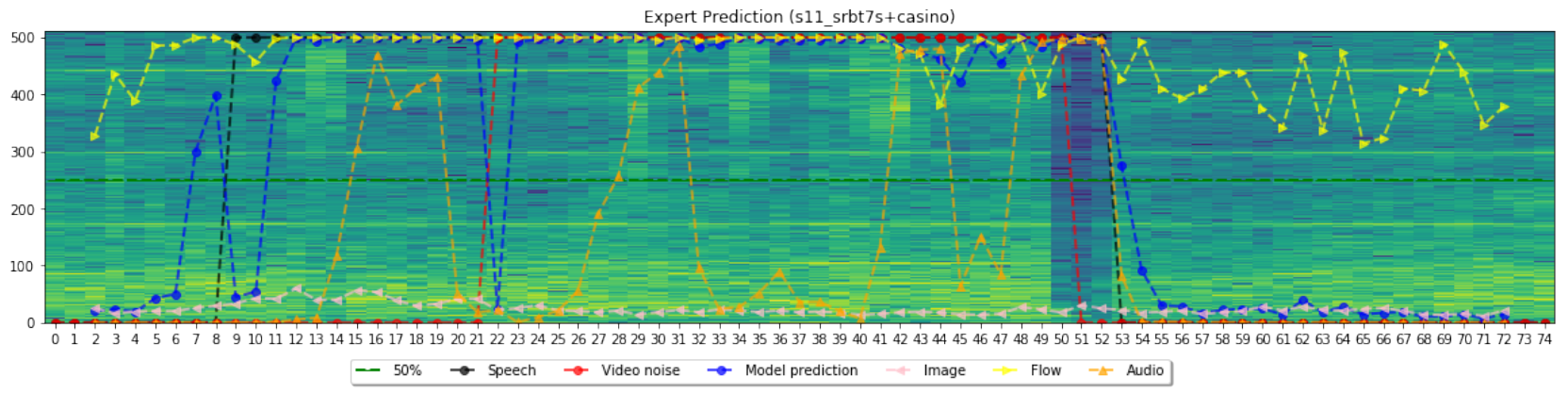}
         \caption{Expert network outputs.}
%         \label{fig:grid-p1}
     \end{subfigure}
     \qquad
     %~ %add desired spacing between images, e. g. ~, \quad, \qquad, \hfill etc. 
       %(or a blank line to force the subfigure onto a new line)
       
     \begin{subfigure}[b]{\textwidth}
         \includegraphics[width=\textwidth]{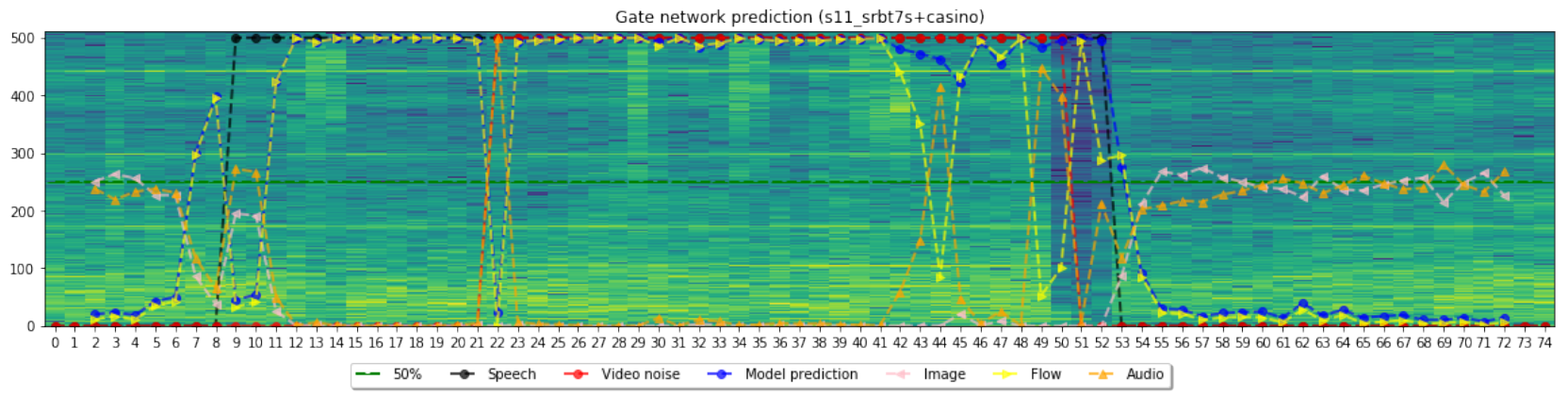}
         \caption{Attention network outputs.}
%         \label{fig:grid-p1-face}
     \end{subfigure}
     \caption{Model outputs (subject 11, speech srbt7s, mixed with casino noise at -5db.}
     \label{fig:p11-srbt7s-casino}
 \end{figure}
 
  \begin{figure}[t!]
     \centering
     \begin{subfigure}[b]{\textwidth}
         \includegraphics[width=\textwidth]{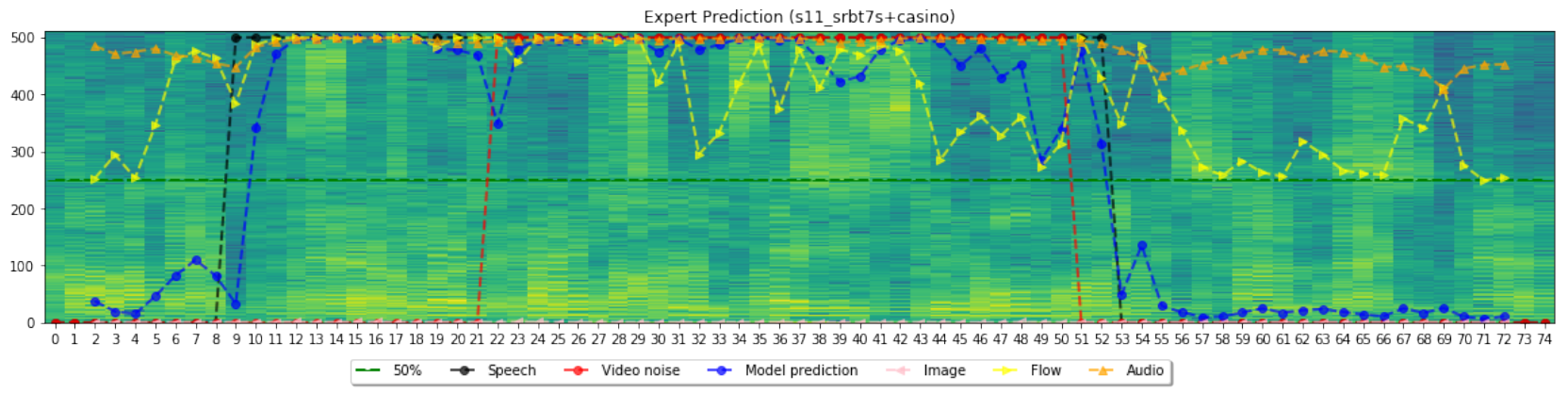}
         \caption{Expert network outputs.}
%         \label{fig:grid-p1}
     \end{subfigure}
     \qquad
     %~ %add desired spacing between images, e. g. ~, \quad, \qquad, \hfill etc. 
       %(or a blank line to force the subfigure onto a new line)
       
     \begin{subfigure}[b]{\textwidth}
         \includegraphics[width=\textwidth]{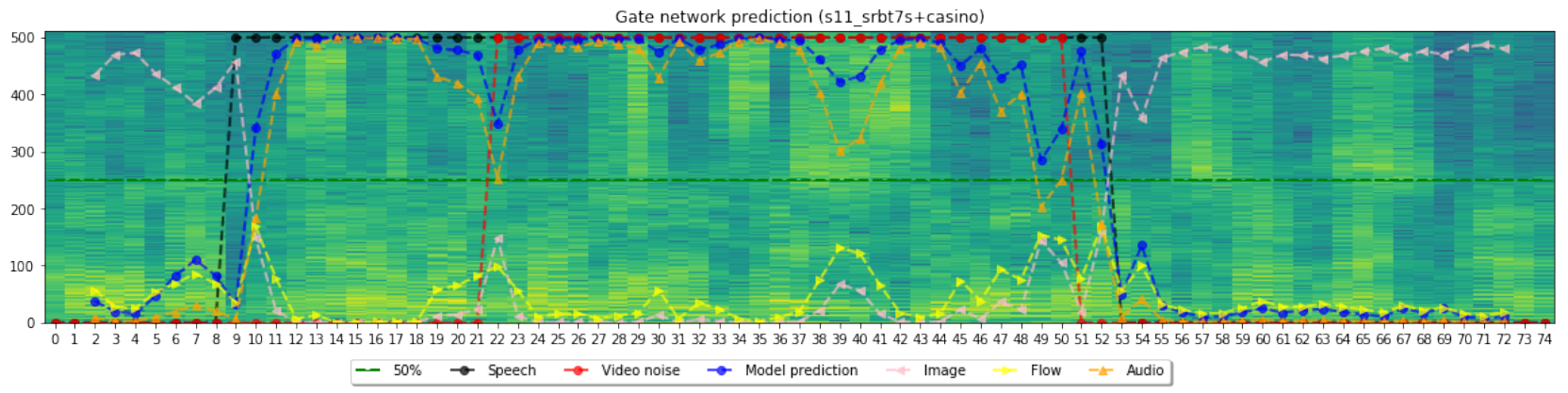}
         \caption{Attention network outputs.}
%         \label{fig:grid-p1-face}
     \end{subfigure}
     \caption{Model outputs (subject 11, speech srbt7s, mixed with 3 random person noise at 0db.)}
     \label{fig:p11-srbt7s-3person}
 \end{figure}

\subsection{Spatial Attention Outputs}
We would like to discuss the spatial attention generated by the gate module. Recall the spatial attention gives the mixing weights to the experts. When a sensor's signal is corrupted, we expect the mixing weight for that sensor would decrease, and the mixing weight for a robust sensor would increase, such that the overall decision is more likely to be correct. In \autoref{fig:p11-srbt7s-3person} we show the attention weights for a speech sample. Notice that speech begins at the 9th frame (black line), and the model prediction lags only one frame. The interesting phenomenon is that at the 22nd frame, the video stream changes to corrupted (red line), causing the mixing weight for audio expert to increase, while the mixing weights for image and optical flow decrease. This suggests that the attention module has learned to identify which sensor inputs are more reliable and assigns mixing weights accordingly. At the 51st frame, the video stream changes to uncorrupted, and we see the mixing weight for image increases to the largest. Recall we have overlap speech audio with noise audio of the entire speech duration, hence video signal is more reliable when uncorrupted.

We notice throughout the entire speech, the attention module prefers to choose only one of the three inputs as the dominant input. It is likely due to the nature of the relationship between the three inputs we have in this particular experiment. Another possibility is the softmax function we used for generating mixing weights may prefer sparse outputs. In order to investigate this property, we can (1) use a different set of sensor inputs, and (2) modify or replace the softmax function to encourage non-sparse outputs.

\section{Conclusion and Future Works}\label{sec:future}
In our speech activity detection experiment, we found that the best multimodal model has $95\%$ accuracy on testing set, whereas for flow model and audio model, the accuracy is $90-92\%$, and $85\%$ for image model. The multimodal model prediction is more accurate and stable than unimodal models. This suggests that multimodal model has successfully combined different sensor data for a common task. We also found the attention function dynamically estimates how reliable each expert is and assigns weights accordingly. Hence the gate module successfully opened the black-box of sensor fusion, which provides insight to the relation between signal and sensor fusion process. The result shows a small step towards a structured sensor fusion method. Speech activity detection involves a sequential binary classification problem. As we discussed, another interesting problem is speech separation (e.g. denoising). We are currently working on this problem.

We found that co-learning using distance-based regularizer decreases prediction accuracy. This is possibly due to co-learning separates modality invariant features and is not designed for speech activity detection. We also proposed to combine probablistic graphical model with deep neural network to construct a model-based sensor fusion model. The model has a co-learning design which tries to separate modality-invariant and modality-dependent features. We are pursuing this direction at the moment. We would like to investigate if model-free and model-based co-learning can successfully separate modality invariant features from modality dependent features. 

The combination of denoising autoendoder, siamese network, and conditional attention mixture model is another work we are currently working on as part of the author's dissertation. Although we choose audio event detection as a application, this framework is general enough to solve other problems, e.g. speaker identification and video event detection in multimodal setting. We will explore these possibilities in future.

%\section{(omit for this preprint) Potential Future Works}
%
%Speech activity detection involves a sequential binary classification problem, which is interesting but may be less challenging to a deep neural network. We have two future directions following current work:
%\begin{enumerate}
%    \item Use more realistic video corruption models.
%    \item Use more adverse face video, e.g. side view. Use more advanced face recognition models.
%    \item Use more advanced face feature extractor.
%    \item Use I3D instead of RNN.
%    \item Online temporal attention: so far all attention models are offline. The attention model has too know the entire sequence. If we have longer training data sequence, we can design a online atte
%    \item Speech enhancement.
%    \item Multimodal automatic speech recognition.
%    \item Other perception problems related to intelligent robot.
%\end{enumerate}

%\section{(omit for this preprint) Regularization Parameter in Variational Inference (complete by end of April)}
%\lijiang{I'm working on this topic for my Learning Theory class (Dr. Khardon). I expect to add this section at the end of April.}
%
%\section{(omit for this preprint) Variational Method for Control and Planning}
%
%\subsection{Variational Methods for Reinforcement Learning}

\section*{Acknowledgement}
Part of this work is based on Lijiang Guo's PhD qualify exam paper. We would like to thank Dr. Geoffrey Fox, Dr. Minje Kim, Dr. Francesco Nesta, Dr. Michael Ryoo and Dr. Lantao Liu for helpful discussions.

%\section*{Appendix}\label{appendix}
%\section*{A. EM solution to conditional mixture of Bernouilli}

\medskip

%
%\begin{figure}[!ht]
%\begin{center}
%% \fbox{\rule{0pt}{2in} \rule{0.9\linewidth}{0pt}}
%   \includegraphics[width=0.8\linewidth]{figures/mixture_local_expert_DNN.png}
%\end{center}
%   \caption{Attention without memory.}
%% \label{fig:long}
%\label{fig:mixture_local_expert_DNN}
%\end{figure}
%
%\begin{figure}[!ht]
%\begin{center}
%% \fbox{\rule{0pt}{2in} \rule{0.9\linewidth}{0pt}}
%   \includegraphics[width=0.8\linewidth]{figures/mixture_local_expert_DNN_colearn.png}
%\end{center}
%   \caption{Regularized attention without memory model.}
%% \label{fig:long}
%\label{fig:mixture_local_expert_DNN_colearn}
%\end{figure}
%
%\begin{figure}[!ht]
%\begin{center}
%% \fbox{\rule{0pt}{2in} \rule{0.9\linewidth}{0pt}}
%   \includegraphics[width=0.8\linewidth]{figures/mixture_local_expert_RNN.png}
%\end{center}
%   \caption{Recurrent attention model.}
%% \label{fig:long}
%\label{fig:mixture_local_expert_RNN}
%\end{figure}

\bibliography{GuoLijiang_lib}
%\bibliography{multimodal}

\end{document}